\documentclass[10pt,twocolumn,twoside]{IEEEtran}

\usepackage[normalem]{ulem}
\usepackage{cite}
\usepackage{url}
\usepackage{epstopdf}
\usepackage{graphicx}
\usepackage{booktabs}
\usepackage{tabularx}
\usepackage{multirow}
\usepackage{subfigure}
\usepackage{amsmath,dsfont}
\usepackage{array}
\usepackage{subfigure}
\usepackage{mathtools}

\usepackage{algorithm,algorithmic}

\usepackage{verbatim}
\usepackage{bbm}

\usepackage{colortbl}
\definecolor{mygray}{gray}{.75}

\usepackage{scalerel,stackengine}

\usepackage{framed}
\usepackage{color}
\usepackage{xcolor}
\usepackage{amssymb}

\newcommand{\argmin}{\operatornamewithlimits{argmin}}
\newcommand{\red}[1]{\textcolor{red}{#1}}
\newcommand{\blue}[1]{\textcolor{blue}{#1}}
\usepackage[normalem]{ulem}

\newcommand{\eg}{\emph{e.g.,}~}
\newcommand{\etal}{\emph{et al.}~}
\newcommand{\ie}{\emph{i.e.,}~}

{\begin{snugshade}\begin{quote}}
{\hfill\end{quote}\end{snugshade}}

\definecolor{shadecolor}{rgb}{0.9,0.9,0.9}

\usepackage[bookmarks=false]{hyperref}

\newcommand{\ourmodel}{Word2VisualVec}

\newcommand{\wordvec}{word2vec}
\newcommand{\videovec}{Word2VideoVec}

\begin{document}
\title{Predicting Visual Features from Text for \\ Image and Video Caption Retrieval}

\author{Jianfeng~Dong,~Xirong~Li, and~Cees~G.~M.~Snoek
\thanks{Manuscript received July 04, 2017; revised January 19, 2018 and March 23, 2018; accepted April 16, 2018. 
This work was supported by NSFC (No. 61672523), the Fundamental Research Funds for the Central Universities, the Research Funds of Renmin University of China (No. 18XNLG19) and the STW STORY project.
The associate editor coordinating the review of this manuscript and approving it for publication was Prof. Benoit HUET.
(\textit{Corresponding author: Xirong Li}).}
\thanks{J. Dong is with the College of Computer Science and Technology, Zhejiang University,
Hangzhou 310027, China (e-mail: danieljf24@zju.edu.cn).}
\thanks{X. Li is with the Key Lab of Data Engineering and Knowledge Engineering, School of Information, Renmin University of China, Beijing 100872, China (e-mail: xirong@ruc.edu.cn).}
\thanks{C. G. M. Snoek is with the Informatics Institute, University of Amsterdam, Amsterdam 1098 XH, The Netherlands (e-mail: cgmsnoek@uva.nl).} 
}

\markboth{IEEE Transactions on Multimedia ,~Vol.~?, No.~?, ?~2018}%
{Dong \MakeLowercase{\textit{et al.}}: Predicting Visual Features from Text for Image and Video Caption Retrieval}

\maketitle

\begin{abstract}
This paper strives to find amidst a set of sentences the one best describing the content of a given image or video. Different from existing works, which rely on a joint subspace for their image and video caption retrieval, we propose to do so in a visual space exclusively. Apart from this conceptual novelty, we contribute \emph{Word2VisualVec}, a deep neural network architecture that learns to predict a visual feature representation from textual input. Example captions are encoded into a textual embedding based on multi-scale sentence vectorization and further transferred into a deep visual feature of choice via a simple multi-layer perceptron. We further generalize Word2VisualVec for video caption retrieval, by predicting from text both 3-D convolutional neural network features as well as a visual-audio representation. Experiments on Flickr8k, Flickr30k, the Microsoft Video Description dataset and the very recent NIST TrecVid challenge for video caption retrieval detail Word2VisualVec's properties, its benefit over textual embeddings, the potential for multimodal query composition and its state-of-the-art results. 
\end{abstract}


\IEEEpeerreviewmaketitle

\section{Introduction} \label{sec:intro}

\IEEEPARstart{T}{his} paper attacks the problem of \emph{image and video caption retrieval}, \ie finding amidst a set of possible sentences the one best describing the content of a given image or video. 
Before the advent of deep learning based approaches to feature extraction, an image or video is typically represented by a bag of quantized local descriptors (known as visual words) while a sentence is represented by a bag of words. These hand-crafted features do not well represent the visual and lingual modalities, and are not directly comparable. Hence, feature transformations are performed on both sides to learn a common latent subspace where the two modalities are better represented and a cross-modal similarity can be computed \cite{rasiwasia2010new,mm2014-autoencoder}.
This tradition continues, as the prevailing image and video caption retrieval methods~\cite{eccv2014-sentence-embedding,iccv15-huawei,cvpr2015-klein-fv,tacl2015-kiros, iclr2016-vendrov,aaai2015-xu-video} prefer to represent the visual and lingual modalities in a common latent subspace.
%
Like others before us {\cite{nips2011-im2text, arxiv2015-knn-caption,acl2015-retrieval-caption}, we consider caption retrieval an important enough problem by itself, and we question the dependence on latent subspace solutions. 
For image retrieval by caption, recent evidence \cite{chami2017amecon} shows that a one-way mapping from the visual to the textual modality outperforms the state-of-the-art subspace based solutions. Our work shares a similar spirit but targets at the opposite direction, \ie image and video caption retrieval.
Our key novelty is that we find the most likely caption for a given image or video by looking for their similarity in the visual feature space exclusively,  as illustrated in Fig. \ref{fig:what}.

\begin{figure}[tb!]
\centering
\includegraphics[width=\columnwidth]{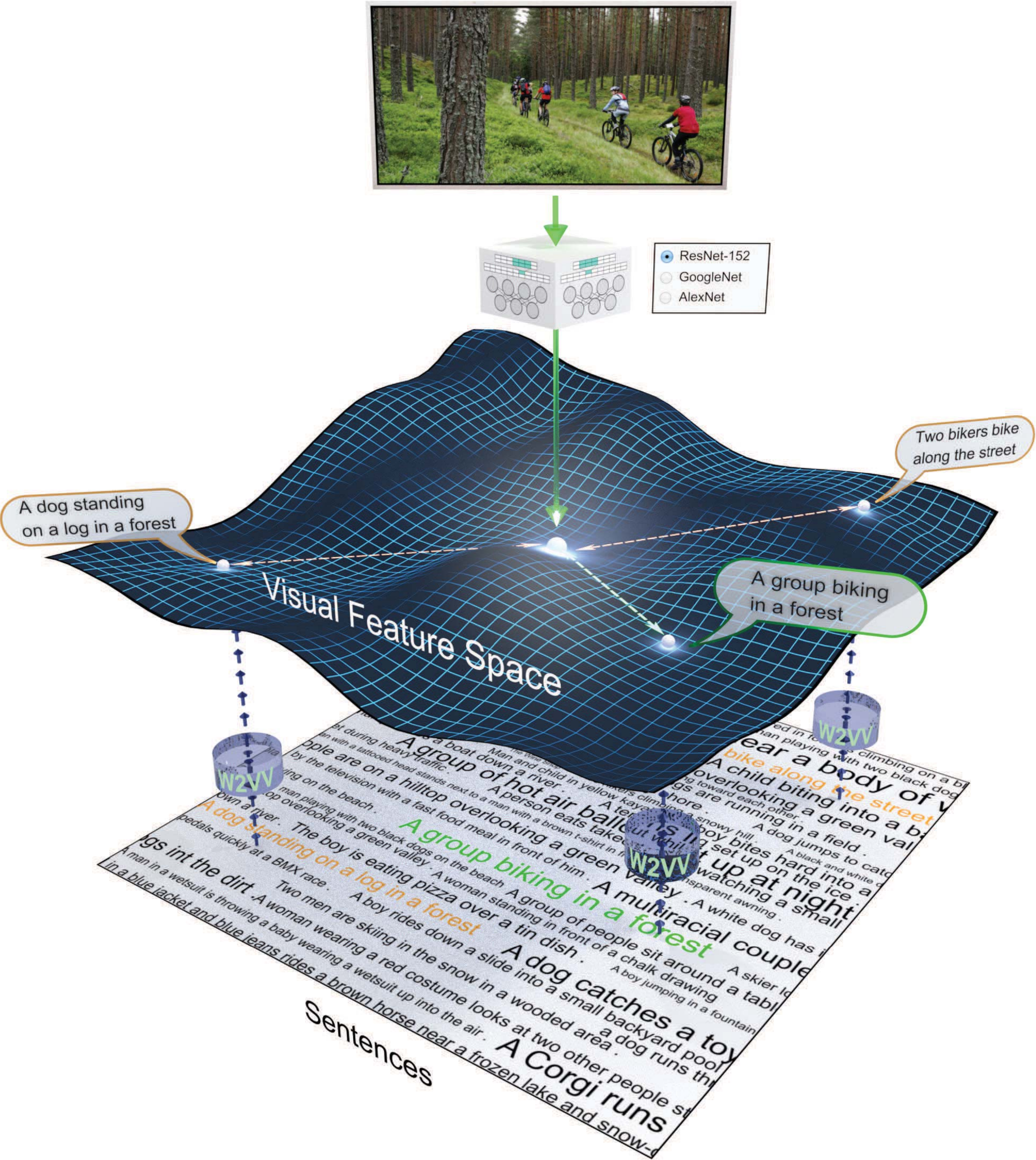}
\caption{\textbf{We propose to perform image and video caption retrieval in a visual feature space exclusively}. This is achieved by Word2VisualVec (W2VV), which predicts visual features from text. As illustrated by the (green) down arrow, a query image is projected into a visual feature space by extracting features from the image content using a pre-trained ConvNet, \eg, GoogleNet or ResNet. As demonstrated by the (black) up arrows, a set of prespecified sentences are projected via W2VV into the same feature space. We hypothesize that the sentence best describing the image content will be the closest to the image in the deep feature space. 
}
\label{fig:what}
\end{figure}

From the visual side we are inspired by the recent progress in predicting images from text \cite{iclr2016-genimage,icml2016-imgsynthesis}. We also depart from the text, but instead of predicting pixels, our model predicts visual features. We consider features from deep convolutional neural networks (ConvNet) \cite{nips2012-hinton,Jia2014Caffe,vggnet,googlenet,cvpr2016-resnet}.  These neural networks learn a textual class prediction for an image by successive layers of convolutions, non-linearities, pooling, and full connections, with the aid of big amounts of labeled images, \eg ImageNet \cite{ILSVRCarxiv14}. Apart from classification, visual features derived from the layers of these networks are superior representations for various challenges in vision \cite{cvpr2014-cnnbaseline,overfeat,ye2015,mm2016-multi,cho2015describing} and multimedia \cite{mm2015jiang-cmu,mm2015-wufei,mmm2016-deep-cmr,mm2016-ngo,hua2016cross}. We also rely on a layered neural network architecture, but rather than predicting a class label for an image, we strive to predict a deep visual feature from a natural language description for the purpose of caption retrieval.

From the lingual side we are inspired by the encouraging progress in sentence encoding by neural language modeling for cross-modal matching \cite{nips13devise,cvpr2015-klein-fv,tacl2014dtrnn,tacl2015-kiros,iclr2016-vendrov,zhang2017cross}.
In particular, \wordvec~\cite{word2vec} pre-trained on large-scale text corpora provides distributed word embeddings, an important prerequisite for vectorizing sentences towards a representation shared with image \cite{nips13devise,cvpr2015-klein-fv} or video  \cite{aaai2015-xu-video,JainICCV15}. In \cite{tacl2015-kiros,iclr2016-vendrov}, a sentence is fed as a word sequence into a recurrent neural network (RNN). The RNN output at the last time step is taken as the sentence feature, which is further projected into a latent subspace. We employ \wordvec~and RNN as part of our sentence encoding strategy as well. What is different is that we continue to transform the encoding into a higher-dimensional visual feature space via a multi-layer perceptron. As we predict visual features from text, we call our approach \emph{Word2VisualVec}.
While both visual and textual modalities are used during training, Word2VisualVec performs a mapping from the textual to the visual modality. Hence, at run time, Word2VisualVec allows the caption retrieval to be performed in the visual space.

We make the following three contributions in this paper: \\
$\bullet$ First, to the best of our knowledge we are the first to solve the caption retrieval problem in the visual space. We consider this counter-tradition approach promising thanks to the effectiveness of deep learning based visual features which are continuously improving. For cross-modal matching, we consider it beneficial to rely on the visual space, instead of a joint space, as it allows us to learn a one-way mapping from natural language text to the visual feature space, rather than a more complicated joint space. \\
$\bullet$ Second, we propose Word2VisualVec to effectively realize the above proposal. Word2VisualVec is a deep neural network based on multi-scale sentence vectorization and a multi-layer perceptron. While its components are known, we consider their combined usage in our overall system novel and effective to transform a natural language sentence into a visual feature vector. We consider prediction of several recent visual features~\cite{Jia2014Caffe,googlenet,cvpr2016-resnet} based on text, but the approach is general and can, in principle, predict any deep visual feature it is trained on. \\ 
$\bullet$ Third, we show how Word2VisualVec can be easily generalized to the video domain, by predicting from text both 3-D convolutional neural network features~\cite{iccv2015-c3d} as well as a visual-audio representation including Mel Frequency Cepstral Coefficients~\cite{opensmile}. Experiments on Flickr8k~\cite{flickr8k}, Flickr30k~\cite{flickr30k}, the Microsoft Video Description dataset~\cite{chen2011collecting} and the very recent NIST TrecVid challenge for video caption retrieval~\cite{AwadTRECVID16} detail Word2VisualVec's properties, its benefit over the word2vec textual embedding, the potential for multimodal query composition and its state-of-the-art results. 

Before detailing our approach, we first highlight in more detail related work.

\section{Related Work} \label{sec:relwork}


\begin{figure*}[tb!]
\centering
\includegraphics[width=2\columnwidth]{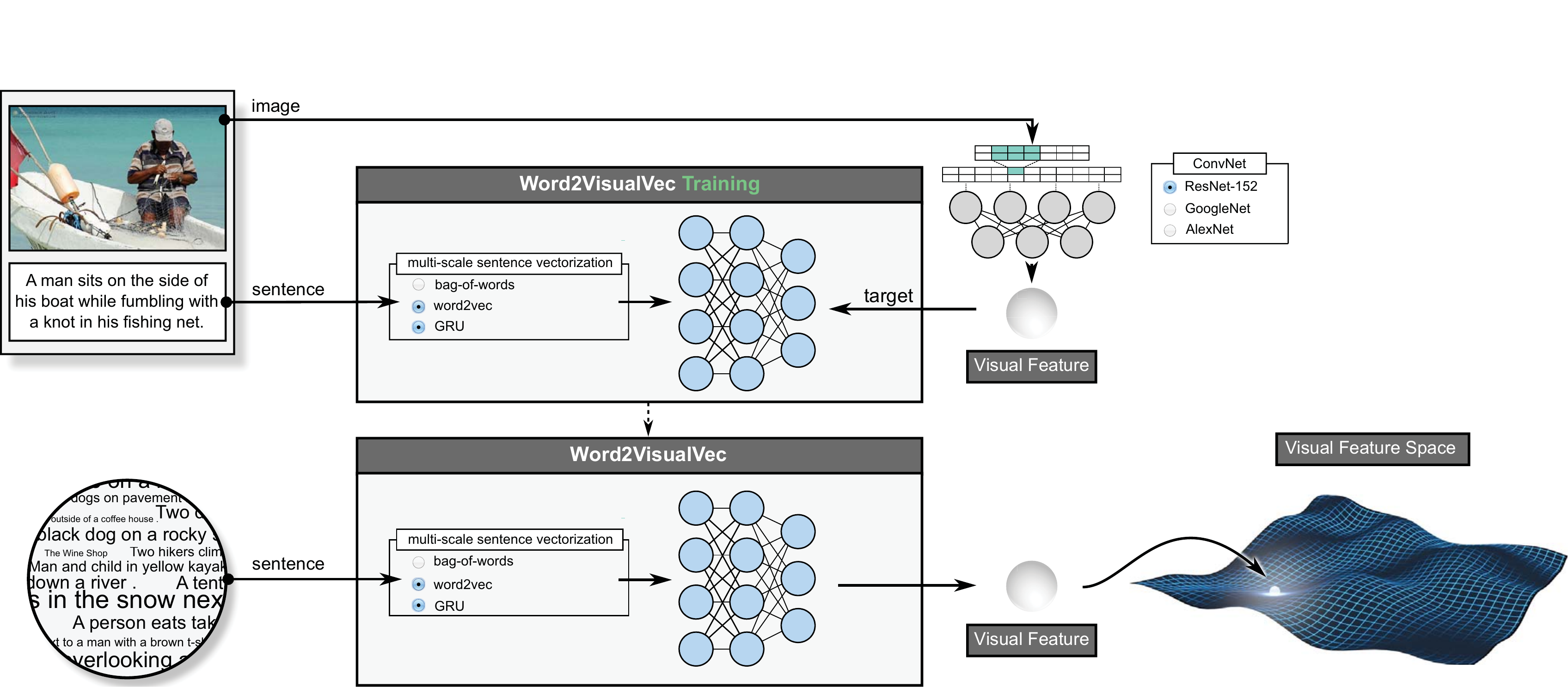}
\caption{\textbf{Word2VisualVec network architecture}. The model first vectorizes an input sentence into a fixed-length vector by relying on bag-of-words, word2vec and a GRU. The vector then goes through a multi-layer perceptron to produce the visual feature vector of choice, from a pre-trained ConvNet such as GoogleNet or ResNet. The network parameters are learned from image-sentence pairs in an end-to-end fashion, with the goal of reconstructing from the input sentence the visual feature vector of the image it is describing.  We rely on the visual feature space for image and video caption retrieval.}
\label{fig:how}
\end{figure*}




\subsection{Caption Retrieval}

Prior to deep visual features, methods for image caption retrieval often resort to relatively complicated models to learn a shared representation to compensate for the deficiency of traditional low-level visual features.  Hodosh \etal \cite{flickr8k} leverage Kernel Canonical Correlation Analysis (CCA), finding a joint embedding by maximizing the correlation between the projected image and text kernel matrices. With deep visual features, we observe an increased use of relatively light embeddings on the image side.
Using the fc6 layer of a pre-trained AlexNet \cite{nips2012-hinton} as the image feature, Gong \etal show that linear CCA compares favorably to its kernel counterpart \cite{eccv2014-sentence-embedding}. Linear CCA is also adopted by Klein \etal \cite{cvpr2015-klein-fv} for visual embedding. More recent models utilize affine transformations to reduce the image feature to a much shorter $h$-dimensional vector, with the transformation optimized in an end-to-end fashion within a deep learning framework \cite{tacl2015-kiros,iclr2016-vendrov,cvpr2016-wang}.
%

%
%

Similar to the image domain, the state-of-the-art methods for video caption retrieval also operate in a shared subspace \cite{aaai2015-xu-video,eccv2016ws-otani,cvpr2017-word-detection}. Xu \etal \cite{aaai2015-xu-video} propose to vectorize each subject-verb-object triplet extracted from a given sentence by a pre-trained \wordvec, and subsequently aggregate the vectors into a sentence-level vector by a recursive neural network. A joint embedding model projects both the sentence vector and the video feature vector, obtained by temporal pooling over frame-level features, into a latent subspace. Otani \etal \cite{eccv2016ws-otani} improve upon \cite{aaai2015-xu-video} by exploiting web image search results of an input sentence, which are deemed helpful for word disambiguation, \eg telling if the word ``keyboard'' refers to a musical instrument or an input device for computers. 
%
%
To learn a common multimodal representation for videos and text, Yu \etal \cite{cvpr2017-word-detection} use two distinct Long Short Term Memory (LSTM) modules to encode the video and text modalities respectively. They then employ a compact bilinear pooling layer to capture implicit interactions between the two modalities.

Different from the existing works, we propose to perform image and video caption retrieval directly in the visual space. This change is important as it allows us to completely remove the learning part from the visual side and focus our energy on learning an effective mapping from natural language text to the visual feature space.


\subsection{Sentence Vectorization}

To convert variably-sized sentences to fixed-sized feature vectors for subsequent learning,  bag-of-words (BoW) is arguably the most popular choice \cite{flickr8k,eccv2014-sentence-embedding,yao2015learning,lei2015predicting}. 
A BoW vocabulary has to be prespecified based on the availability of words describing the training images. As collecting image-sentence pairs at a large-scale is both labor intensive and time consuming, the amount of words covered by BoW is bounded.
To overcome this limit, a distributional text embedding provided by \wordvec~\cite{word2vec} is gaining increased attention. 
The word embedding matrix used in \cite{nips13devise,aaai2015-xu-video,eccv2016ws-otani,mm2016-robust} is instantiated by a \wordvec~model pre-trained on large-scale text corpora. In Frome \etal \cite{nips13devise}, for instance, the input text is vectorized by averaging the \wordvec~vectors of its words. Such a mean pooling strategy results in a dense representation that could be less discriminative than the initial BoW feature. As an alternative, Klein \etal \cite{cvpr2015-klein-fv} and their follow-up \cite{cvpr2016-wang}  perform fisher vector pooling over word vectors. 

Beside BoW and word2vec, we observe an increased use of RNN-based sentence vectorization. Socher \etal design a Dependency-Tree RNN that learns vector representations for sentences based on their dependency trees~\cite{tacl2014dtrnn}. Lev \etal \cite{eccv2016-rnn_fv} propose RNN fisher vectors on the basis of \cite{cvpr2015-klein-fv}, replacing the Gaussian model by a RNN model that takes into account the order of elements in the sequence.
Kiros \etal  \cite{tacl2015-kiros} employ  an LSTM to encode a sentence, using the LSTM's hidden state at the last time step as the sentence feature. In a follow-up work, Vendrov \etal replace LSTM by a Gated Recurrent Unit (GRU) which has less parameters to tune \cite{iclr2016-vendrov}. While RNN and its LSTM or GRU variants have demonstrated promising results for generating visual descriptions \cite{vinyals2015show,iclr15-ma-mrnn, mm16-bilstm, mm2016-early}, they tend to be over-sensitive to word orders by design. Indeed Socher \etal \cite{tacl2014dtrnn} suggest that for caption retrieval, models invariant to surface changes, such as word order, perform better.


In order to jointly exploit the merits of the BoW, word2vec and RNN based representations, we consider in this paper multi-scale sentence vectorization.  Ma \etal \cite{iccv15-huawei} have made a first attempt in this direction. In their approach three multimodal ConvNets are trained on feature maps, formed by merging the image embedding vector with word, phrase and sentence embedding vectors. The relevance between an image and a sentence is estimated by late fusion of the individual matching scores. By contrast, we perform multi-scale sentence vectorization in an early stage, by merging BoW, word2vec and GRU sentence features and letting the model figure out the optimal way for combining them. 
Moreover, at run time the multi-modal network by \cite{iccv15-huawei} requires a query image to be paired with each of the test sentences as the network input. By contrast, our Word2VisualVec model predicts visual features from text alone, meaning the vectorization can be precomputed.  An advantageous property for caption retrieval on large-scale image and video datasets.
\section{Word2VisualVec} \label{sec:approach}

We propose to learn a mapping that projects a natural language description into a visual feature space. Consequently, the relevance between a given visual instance $x$ and a specific sentence $q$ can be directly computed in this space.
More formally, let $\phi(x) \in \mathbb{R}^d$ be a $d$-dimensional visual feature vector. A pretrained ConvNet, apart from its original mission of visual class recognition, has now been recognized as an effective visual feature extractor \cite{cvpr2014-cnnbaseline}. We follow this good practice, instantiating $\phi(x)$ with a ConvNet feature vector. 
We aim for a sentence representation $r(q)\in \mathbb{R}^d$ such that the similarity can be expressed by the cosine similarity between $\phi(x)$ and $r(q)$.
The mapping is optimized by minimizing the Mean Squared Error between the vector of a training sentence and the vector of the visual instance the sentence is describing.
The proposed mapping model \ourmodel~is designed to produce $r(q)$, as visualized in Fig. \ref{fig:how} and detailed next.

%

\subsection{Architecture} \label{sssec:arch}

\textbf{Multi-scale sentence vectorization}. 
To handle sentences of varying length, we choose to first vectorize each sentence. We propose multi-scale sentence vectorization that utilizes BoW, word2vec and RNN based text encodings.

BoW is a classical text encoding method. Each dimension in a BoW vector corresponds to the occurrence of a specific word in the input sentence, \ie
\begin{equation}
s_{bow}(q) = (c(w_1,q), c(w_2,q),\ldots,c(w_m,q)),
\end{equation}
where $c(w,q)$ returns the occurrence of word $w$ in $q$, and $m$ is the size of a prespecified vocabulary. 
A drawback of Bow is that its vocabulary is bounded by the words used in the multi-modal training data, which is at a relatively small scale compared to a text corpus containing millions of words.
Given \textit{faucet} as a novel word, for example, ``A little girl plays with a faucet'' will not have the main object encoded in its BoW vector. 
Notice that setting a large vocabulary for BoW is unhelpful, as words without training images will always have zero value and thus will not be effectively modeled.
To compensate for such a loss, we further leverage word2vec. By learning from a large-scale text corpus, the vocabulary of word2vec is much larger than its BoW counterpart. We obtain the embedding vector of the sentence by mean pooling over its words, \ie
\begin{equation}
s_{word2vec}(q) :=\frac{1}{|q|}\sum_{w \in q} v(w),
\end{equation}
where $v(w)$ denotes individual word embedding vectors, $|q|$ is the sentence length. Previous works employ \wordvec~trained on web documents as their word embedding matrix \cite{nips13devise,vinyals2015show,iccv15-huawei}. However, recent studies suggest that \wordvec~trained on Flickr tags better captures visual relationships than its counterpart learned from web documents \cite{li2015zero,mm2015emoji}. We therefore train a 500-dimensional \wordvec~model on English tags of 30 million Flickr images, using the skip-gram algorithm \cite{word2vec}. 
This results in a vocabulary of 1.7 million words.

Despite their effectiveness, the BoW and word2vec representations ignore word orders in the input sentence. As such, they cannot discriminate between ``a dog follows a person'' and ``a person follows a dog''. To tackle this downside, we employ an RNN, which is known to be effective for modeling long-term word dependency in natural language text. In particular, we adopt a GRU \cite{cho2014learning}, which has less parameters than LSTM and presumably requires less amounts of training data. At a specific time step $t$, let $v_t$ be the embedding vector of the $t$-th word, obtained by performing a lookup on a word embedding matrix $W_e$.  GRU receives inputs from  $v_t$ and the previous hidden state $h_{t-1}$, and accordingly the new hidden state $h_t$ is updated as follows,
\begin{equation}
\begin{array}{l}
z_{t} = \sigma (W_{zv}v_{t} + W_{zh}h_{t-1} + b_{z}), \\
r_{t} = \sigma (W_{rv}v_{t} + W_{rh}h_{t-1} + b_{r}),  \\
\widetilde{h_{t}} = tanh (W_{hv}v_{t} + W_{hh}(r_{t}\odot  h_{t-1}) + b_{h}),  \\
h_{t} = (1-z_{t})\odot h_{t-1} + z_{t} \odot  \widetilde{h_{t}},
\end{array} 
\label{eq:gru}
\end{equation}
where $z_{t}$ and $r_{t}$ denote the update and reset gates at time $t$ respectively, while $W$ and $b$ with specific subscripts are weights and bias parameterizing the corresponding gates. The symbol $\odot$ indicates element-wise multiplication, while $\sigma(\cdot)$ is the sigmoid activation function. We re-use word2vec previously trained on the Flickr tags to initialize $W_e$. The last hidden state $h_{|q|}$ is taken as the RNN based representation of the sentence.

Multi-scale sentence vectorization is obtained by concatenating the three representations, that is
\begin{equation}
s(q) =  [s_{bow}(q), s_{word2vec}(q), h_{|q|}].
\end{equation}

\textbf{Text transformation via a multilayer perceptron}. 
The sentence vector $s(q)$ goes through subsequent hidden layers until it reaches the output layer $r(q)$, which resides in the visual feature space. More concretely, by applying an affine transformation on $s(q)$, followed by an element-wise ReLU activation $\sigma(z) = \max(0,z)$, we obtain the first hidden layer $h_1(q)$ of an $l$-layer \ourmodel~as: 
\begin{equation}
h_1(q)=\sigma(W_{1}  s(q)+b_{1}).
\end{equation}
The following hidden layers are expressed by:
\begin{equation}
h_i(q)=\sigma(W_{i} h_{i-1}(q)+b_{i}), i = 2,...,l-2,
\end{equation}
where $W_{i}$ parameterizes the affine transformation of the $i$-th hidden layer and $b_{i}$ is a bias terms. In a similar manner, we compute the output layer $r(q)$ as: 
\begin{equation}
r(q)=\sigma(W_l  h_{l-1}(q)+b_l).
\end{equation}
Putting it all together, the learnable parameters are represented by 
$\theta=[W_e, W_{z.}, W_{r.}, W_{h.}, b_z, b_r, b_h, W_1, b_1, \ldots, W_l, b_l]$.
                                                         
In principle, the learning capacity of our model grows as more layers are used. This also means more solutions exist which minimize the training loss, yet are suboptimal for unseen test data.  We analyze in the experiments how deep \ourmodel~can go without losing its generalization ability.

\subsection{Learning algorithm}

\textbf{Objective function}. 
For a given image, different persons might describe the same visual content with different words. For example,  ``A dog leaps over a log'' versus ``A dog is leaping over a fallen tree''. The verb leap in different tenses essentially describe the same action, while a log and a fallen tree can have similar visual appearance. Projecting the two sentences into the same visual feature space has the effect of implicitly finding such correlations. In order to reconstruct the visual feature $\phi(x)$ directly from $q$, we use Mean Squared Error (MSE) as our objective function. We have also experimented with the marginal ranking loss, as commonly used in previous works \cite{grangier2008discriminative,Bai2009Polynomial,nips13devise,cvpr2015-neuraltalk}, but found MSE yields better performance.

The MSE loss $l_{mse}$ for a given training pair is defined as: 
\begin{equation} \label{eq:loss_mse}
l_{mse}(x,q ;\theta) =  (r(q)-\phi(x))^2.
\end{equation}
We train \ourmodel~to minimize the overall MSE loss on a given training set $\mathcal{D} = \{(x,q)\}$, containing a number of relevant image-sentence pairs:
\begin{equation}\label{eq:obj_mse}
\argmin_{\theta} \sum_{(x,q)\in \mathcal{D}}  l_{mse}(x,q ;\theta).
\end{equation}


\textbf{Optimization}.
We solve Eq. (\ref{eq:obj_mse}) using stochastic gradient descent with RMSprop \cite{tijmen2012rmsprop}. This optimization algorithm divides the learning rate by an exponentially decaying average of squared gradients, to prevent the learning rate from effectively shrinking over time.
We empirically set the initial learning rate $\eta=0.0001$, decay weights $\gamma=0.9$ and small constant $\epsilon=10^{-6}$ for RMSprop. We apply dropout to all hidden layers in \ourmodel~to mitigate model overfitting. Lastly, we take an empirical learning schedule as follows.
Once the validation performance does not increase in three consecutive epochs, we divide the learning rate by 2. Early stop occurs if the validation performance does not improve in ten consecutive epochs.  The maximal number of epochs is 100.

\subsection{Image Caption Retrieval}

For a given image, we select from a given sentence pool the sentence deemed most relevant with respect to the image. Note that image-sentence pairs are required only for training \ourmodel. For a test sentence, its $r(q$) is obtained by forward computation through the \ourmodel~network, without the need of any test image. Hence, the sentence pool can be vectorized in advance. Image caption retrieval in our case boils down to finding the sentence nearest to the given image in the visual feature space. We use the cosine similarity between $r(q$) and  the image feature $\phi(x)$,  as this similarity normalizes feature vectors and is found to be better than the dot product or mean square error according to our preliminary experiments.

\begin{table*} [tb!]
\renewcommand{\arraystretch}{1.1}
\caption{\textbf{Performance of image caption retrieval on Flickr8k.}
Multi-scale sentence vectorization combined with the ResNet-152 feature is the best.
}
\label{tab:exp-flickr8k}
\centering 
\scalebox{1.0}{
\begin{tabular}{@{}l r@{\hskip 0.15in} rrr r@{\hskip 0.15in}  rrr r@{\hskip 0.15in} rrr r@{\hskip 0.15in} rrr @{}}
\toprule
 & & \multicolumn{3}{c}{\textbf{BoW}} & & \multicolumn{3}{c}{\textbf{word2vec}} & & \multicolumn{3}{c}{\textbf{GRU}} & & \multicolumn{3}{c}{\textbf{Multi-scale}}\\
\cmidrule(l){3-5}  \cmidrule(l){7-9} \cmidrule(l){11-13} \cmidrule(l){15-17}
\textbf{Visual Features}   & & R@1 & R@5 & R@10  &  & R@1 & R@5 & R@10  &  & R@1 & R@5 & R@10 &  & R@1 & R@5 & R@10\\
\cmidrule{1-17}
CaffeNet           & & 20.7 & 43.3 & 55.2    &  & 18.9 & 42.3 & 54.2 &   & 21.2 & 44.7 & 56.1  & & 23.1 & 47.1 & 57.7 \\
GoogLeNet          & & 27.1 & 53.5 & 64.9    &  & 24.7 & 51.6 & 64.1 &   & 25.1 & 51.9 & 64.2  & & 28.8 & 54.5 & 68.2 \\ 
GoogLeNet-shuffle  & & 32.2 & 57.4 & 72.0    &  & 30.2 & 57.6 & 70.5 &   & 32.9 & 59.5 & 70.5  & & 35.4 & 63.1 & 74.0 \\ 
ResNet-152         & & 34.7 & 62.9 & 74.7    &  & 32.1 & 62.9 & 75.5 &   & 33.4 & 63.1 & 75.3  & & \textbf{36.3} & \textbf{66.4} & \textbf{78.2} \\ 
\bottomrule
\end{tabular}
 }
\end{table*}

\begin{table*} [tb!]
\renewcommand{\arraystretch}{1.1}
\caption{\textbf{Performance of image caption retrieval on Flickr30k.}
Multi-scale sentence vectorization combined with the ResNet-152 feature is the best.
}
\label{tab:exp-flickr30k}
\centering 
\scalebox{1.0}{
\begin{tabular}{@{}l r@{\hskip 0.15in} rrr r@{\hskip 0.15in}  rrr r@{\hskip 0.15in} rrr r@{\hskip 0.15in} rrr @{}}
\toprule
 & & \multicolumn{3}{c}{\textbf{BoW}} & & \multicolumn{3}{c}{\textbf{word2vec}} & & \multicolumn{3}{c}{\textbf{GRU}} & & \multicolumn{3}{c}{\textbf{Multi-scale}}\\
\cmidrule(l){3-5}  \cmidrule(l){7-9} \cmidrule(l){11-13} \cmidrule(l){15-17}
\textbf{Visual Features}   & & R@1 & R@5 & R@10  &  & R@1 & R@5 & R@10  &  & R@1 & R@5 & R@10 &  & R@1 & R@5 & R@10\\
\cmidrule{1-17}
CaffeNet           & & 24.4 & 47.1 & 57.1   &  & 18.9 & 42.3 & 54.2  &   & 24.1 & 46.4 & 57.4   & & 24.9 & 50.4 & 60.8 \\
GoogLeNet          & & 32.2 & 58.3 & 67.7   &  & 24.7 & 51.6 & 64.1  &   & 33.6 & 56.8 & 67.2   & & 33.9 & 62.2 & 70.8  \\ 
GoogLeNet-shuffle  & & 38.6 & 66.4 & 75.2   &  & 30.2 & 57.6 & 70.5  &   & 38.6 & 64.8 & 76.7   & & 41.3 & 69.1 & 78.6 \\ 
ResNet-152         & & 41.8 & 70.9 & 78.6   &  & 36.5 & 65.0 & 75.1  &   & 42.0 & 70.4 & 80.1   & & \textbf{45.9} & \textbf{71.9} & \textbf{81.3}   \\ 
\bottomrule
\end{tabular}
 }
\end{table*}

\begin{table*} [tb!]
\renewcommand{\arraystretch}{1.1}
\caption{\textbf{Performance of video caption retrieval on MSVD.}
Multi-scale sentence vectorization combined with the GoogLeNet-shuffle feature is the best.
}
\label{tab:exp-msvd}
\centering 
\scalebox{1.0}{
\begin{tabular}{@{}l r@{\hskip 0.15in} rrr r@{\hskip 0.15in}  rrr r@{\hskip 0.15in} rrr r@{\hskip 0.15in} rrr @{}}
\toprule
 & & \multicolumn{3}{c}{\textbf{BoW}} & & \multicolumn{3}{c}{\textbf{word2vec}} & & \multicolumn{3}{c}{\textbf{GRU}} & & \multicolumn{3}{c}{\textbf{Multi-scale}}\\
\cmidrule(l){3-5}  \cmidrule(l){7-9} \cmidrule(l){11-13} \cmidrule(l){15-17}
\textbf{Visual Features}   & & R@1 & R@5 & R@10  &  & R@1 & R@5 & R@10  &  & R@1 & R@5 & R@10 &  & R@1 & R@5 & R@10\\
\cmidrule{1-17}
CaffeNet           & & 9.4  & 19.9 & 26.7    &  & 8.7  & 22.2 & 31.3  &   & 9.6  & 19.4 & 26.9  & & 9.6  & 21.9 & 30.6 \\
GoogLeNet          & & 14.2 & 27.5 & 36.0    &  & 14.5 & 30.3 & 39.7  &   & 16.0 & 33.1 & 43.0  & & 17.2 & 33.7 & 42.8 \\ 
GoogLeNet-shuffle  & & 14.8 & 29.6 & 37.2    &  & 16.6 & 33.7 & 43.4  &   & 16.6 & 35.1 & 42.8  & & \textbf{18.5} & \textbf{36.7} & 45.1 \\ 
ResNet-152         & & 15.8 & 32.1 & 39.9    &  & 16.4 & 34.8 & \textbf{46.6}  &   & 15.8 & 31.3 & 41.8  & & 16.1 & 34.5 & 43.1 \\ 
C3D                & & 10.4 & 22.5 & 28.4    &  & 14.8 & 34.5 & 44.0  &   & 13.1 & 26.6 & 33.4  & & 14.9 & 27.8 & 35.5 \\ 
\bottomrule
\end{tabular}
 }
\end{table*}

\subsection{Video Caption Retrieval} \label{sec:w2vv-video}
\ourmodel~is also applicable for video as long as we have an effective vectorized representation of video. Again, different from previous methods for video caption retrieval that execute in a joint subspace \cite{aaai2015-xu-video,eccv2016ws-otani}, we project sentences into the video feature space.

Following the good practice of using pre-trained ConvNets for video content analysis \cite{naacl2015-video-to-text,ye2015,mettes2016imagenet,pan2016jointly}, we extract features by applying image ConvNets on individual frames and 3-D ConvNets \cite{iccv2015-c3d} on consecutive-frame sequences. For short video clips, as used in our experiments, mean pooling over video frames is considered reasonable \cite{naacl2015-video-to-text,pan2016jointly}. Hence, the visual feature vector of each video is obtained by averaging the feature vectors of its frames.  Note that longer videos open up possibilities for further improvement of Word2VisualVec by exploiting temporal order of video frames, \eg \cite{cvpr2016-yu-video}.  The audio channel of a video sometime provides complementary information to the visual channel. For instance, to help decide whether a person is talking or singing. To exploit this channel, we extract a bag of quantized Mel-frequency Cepstral Coefficients (MFCC)  \cite{opensmile} and concatenate it with the previous visual feature. \ourmodel~is trained to predict such a visual-audio feature, as a whole, from input text.

\ourmodel~is used in a principled manner, transforming an input sentence to a video feature vector, let it be visual or visual-audio. For the sake of clarity we term the video variant \textit{Word2VideoVec}.

\section{Experiments} \label{sec:eval}


\subsection{Properties of Word2VisualVec} \label{ssec:eval-design-choice}

We first investigate the impact of major design choices, \eg \textit{how to vectorize an input sentence?}.
Before detailing the investigation, we first introduce data and evaluation protocol.

\textbf{Data}.  
For image caption retrieval, we use two popular benchmark sets, Flickr8k \cite{flickr8k} and Flickr30k \cite{flickr30k}. Each image is associated with five crowd-sourced English sentences, which briefly describe the main objects and scenes present in the image. For video caption retrieval we rely on the Microsoft Video Description dataset (MSVD) \cite{chen2011collecting}.  Each video is labeled with 40 English sentences on average. The videos are short, usually less than 10 seconds long.
For the ease of cross-paper comparison, we follow the identical data partitions as used in \cite{cvpr2015-neuraltalk,cvpr2015-klein-fv,iclr2016-vendrov} for images and \cite{naacl2015-video-to-text} for videos. 
That is, training / validation / test is 6k / 1k / 1k for Flickr8k, 29K / 1,014 / 1k for Flickr30k, 
and 1,200 / 100 / 670 for MSVD.

\textbf{Visual features}.  
A deep visual feature is determined by a specific ConvNet and its layers. We experiment with four pretrained 2-D ConvNets, \ie CaffeNet \cite{Jia2014Caffe}, GoogLeNet \cite{googlenet},  GoogLeNet-shuffle \cite{mettes2016imagenet} and ResNet-152 \cite{cvpr2016-resnet}. 
The first three 2-D ConvNets were trained using images containing 1K different visual objects as defined in the Large Scale Visual Recognition Challenge \cite{ILSVRCarxiv14}. GoogLeNet-shuffle follows GoogLeNet's architecture, but is re-trained using a bottom-up reorganization of the complete 22K ImageNet hierarchy, excluding over-specific classes and classes with few images and thus making the final classes more balanced.
For the video dataset, we further experiment with a 3-D ConvNet \cite{iccv2015-c3d}, trained on one million sports videos containing 487 sport-related concepts \cite{karpathy2014large}.
As the videos were muted, we cannot evaluate Word2VideoVec with audio features. 
We tried multiple layers of each ConvNet model and report the best performing layer.
Finally we use the fc7 layer for CaffeNet (4,096-dim), the pool5 layer for GoogleNet (1,024-dim), GoogleNet-shuffle (1,024-dim) and ResNet-152 (2,048-dim), and the fc6 layer for C3D (4,096-dim). 

\textbf{Details of the model}.
The size of the word2vec and GRU layers is 500 and 1,024, respectively. The size of the BoW layer depends on training data, which is 2,535, 7,379 and 3,030 for Flickr8k, Flickr30k and MSVD, respectively (with words appearing less than five times in the corresponding training set removed). Accordingly, the size of the composite vectorization layer is 4,059, 8,903 and 4,554, respectively. The size of the hidden layers is 2,048. The number of layers is three unless otherwise stated. Code is available at \url{https://github.com/danieljf24/w2vv}.

\textbf{Evaluation protocol}. 
The training, validation and test set are used for model training, model selection and performance evaluation, respectively, and exclusively. For performance evaluation, each test caption is first vectorized by a trained Word2VisualVec. Given a test image/video query, we then rank all the test captions in terms of their similarities with the image/video query in the visual feature space. The performance is evaluated based on the caption ranking.
Following the common convention \cite{flickr8k,iccv15-huawei,iclr2016-vendrov}, we report rank-based performance metrics $R@K$ ($K = 1, 5, 10$). $R@K$ computes the percentage of test images for which at least one correct result is found among the  top-$K$ retrieved sentences. Hence, higher $R@K$ means better performance.

\textbf{How to vectorize an input sentence?}
As shown in Table \ref{tab:exp-flickr8k}, \ref{tab:exp-flickr30k} and \ref{tab:exp-msvd}, multi-scale sentence vectorization outperforms its single-scale counterparts.
Table \ref{tab:opt_text_encode} shows examples for which a particular vectorization method is particularly suited.  In the first two rows, word2vec performs better than BoW and GRU, because the main words \textit{rottweiler} and \textit{quad}  are not in the vocabularies of BoW and GRU. However, the use of word2vec sometimes has the side effect of overweighting high-level semantic similarity between words. E.g., \textit{beagle} in the third row is found to be closer to \textit{dog}  than to \textit{hound}, and \textit{woman} in the fourth row is found to be more close to \textit{man} than to \textit{lady} in the word2vec space. In this case, the resultant Word2VisualVec vector is less discriminative than its BoW counterpart. Since GRU is good at modeling long-term word dependency, it performs the best in the last two rows, where the captions are more narrative.

\begin{table} [tb!]
\renewcommand{\arraystretch}{1.2}
\caption{\textbf{Caption ranks by Word2VisualVec with distinct sentence vectorization strategies}. Lower rank means better performance.}
\label{tab:opt_text_encode}
\centering 
\scalebox{1.0}{
\begin{tabular}{@{}l p{6cm}}
\toprule
%
\textbf{Query image}   & \textbf{Ground-truth caption and its ranks} \\
\cmidrule{1-2}
\multirow{3}{*}{\includegraphics[width=1.8cm,height=1.8cm]{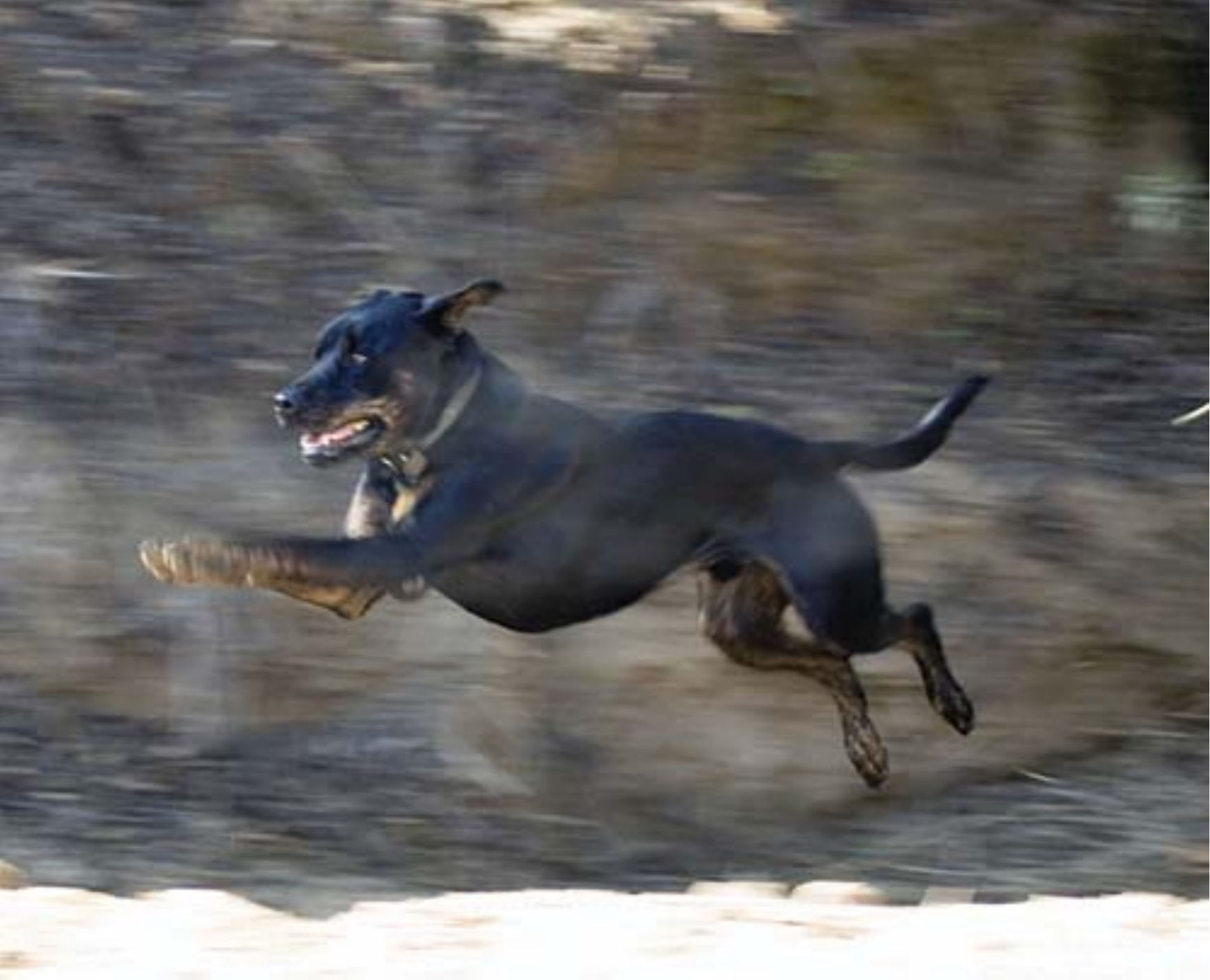}}   & \\
& A rottweiler running.    \\
& BoW$\rightarrow$857  \qquad  \textbf{word2vec$\rightarrow$84} \qquad  GRU$\rightarrow$841\\
& \\
& \\ [4pt]
\multirow{3}{*}{\includegraphics[width=1.8cm,height=1.8cm]{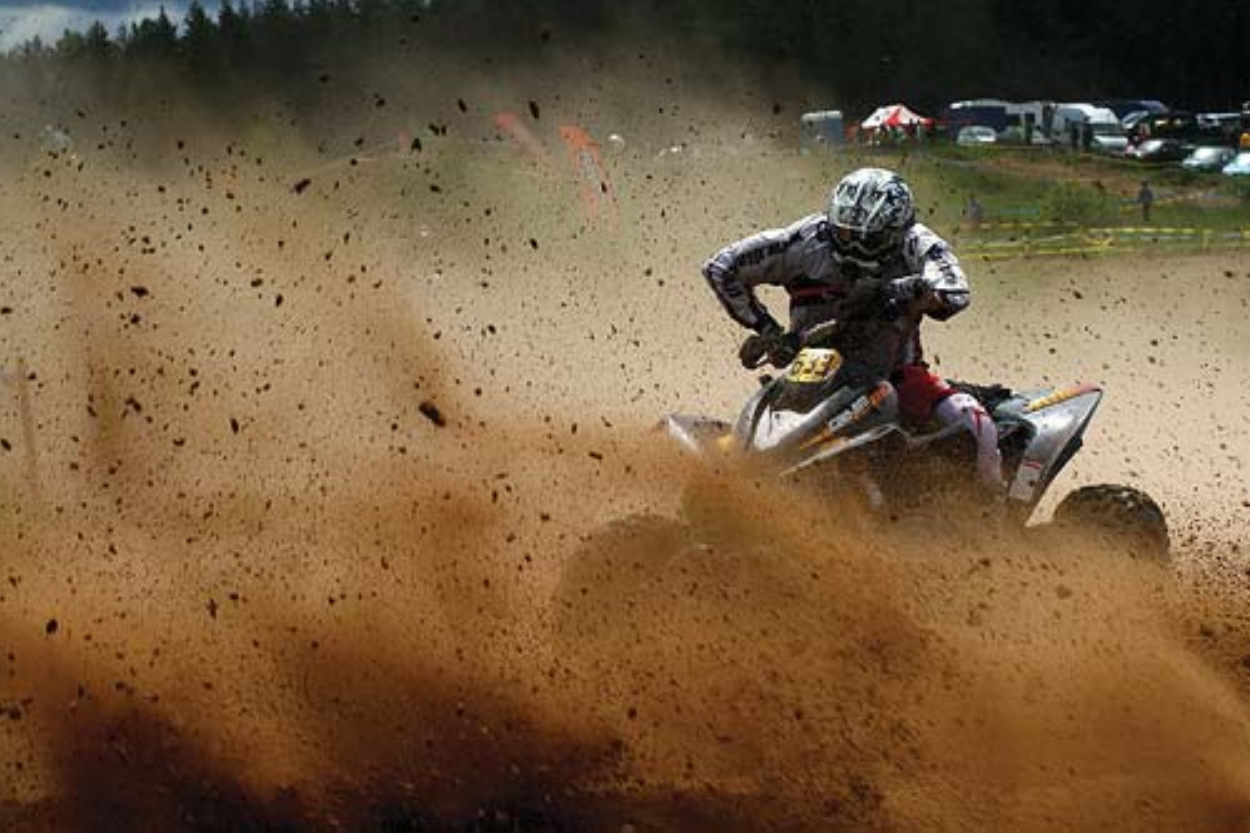}}   & \\
& A quad sends dirt flying into the air.   \\
& BoW$\rightarrow$41  \qquad  \textbf{word2vec$\rightarrow$5} \qquad  GRU$\rightarrow$28\\
& \\
& \\ [4pt]
\multirow{4}{*}{\includegraphics[width=1.8cm,height=1.8cm]{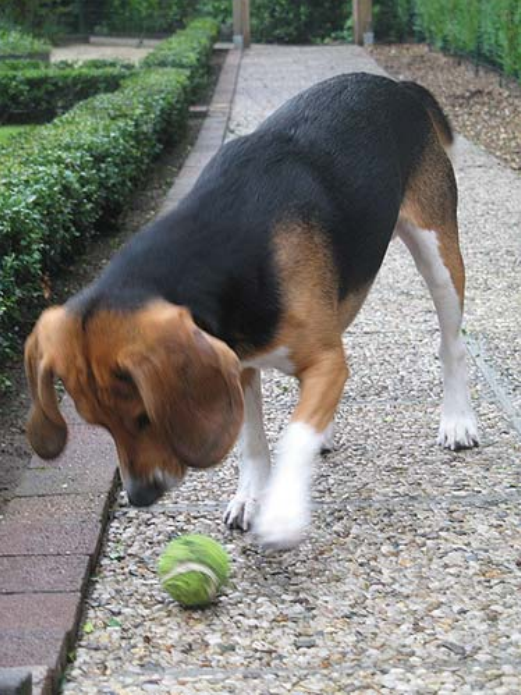}}   & \\
& A white-footed beagle plays with a tennis ball on a garden path.    \\
& \textbf{BoW$\rightarrow$7}  \qquad  word2vec$\rightarrow$22 \qquad  GRU$\rightarrow$65\\
& \\ [4pt]
\multirow{4}{*}{\includegraphics[width=1.8cm,height=1.8cm]{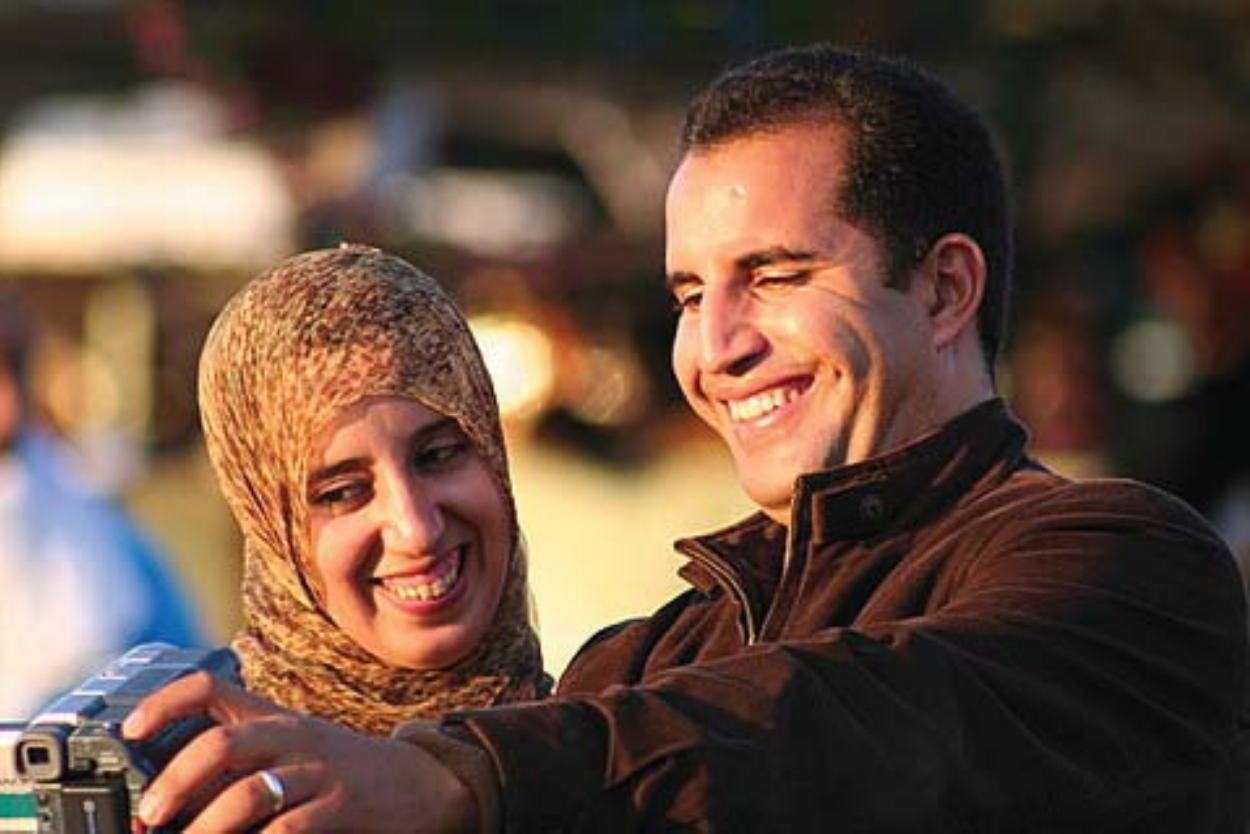}}   & \\
& A man in a brown sweater and a woman smile for their video camera.    \\
& \textbf{BoW$\rightarrow$3}  \qquad  word2vec$\rightarrow$43 \qquad  GRU$\rightarrow$16\\
& \\ [4pt]
\multirow{4}{*}{\raisebox{-0.5\totalheight}{\includegraphics[width=1.8cm,height=1.8cm]{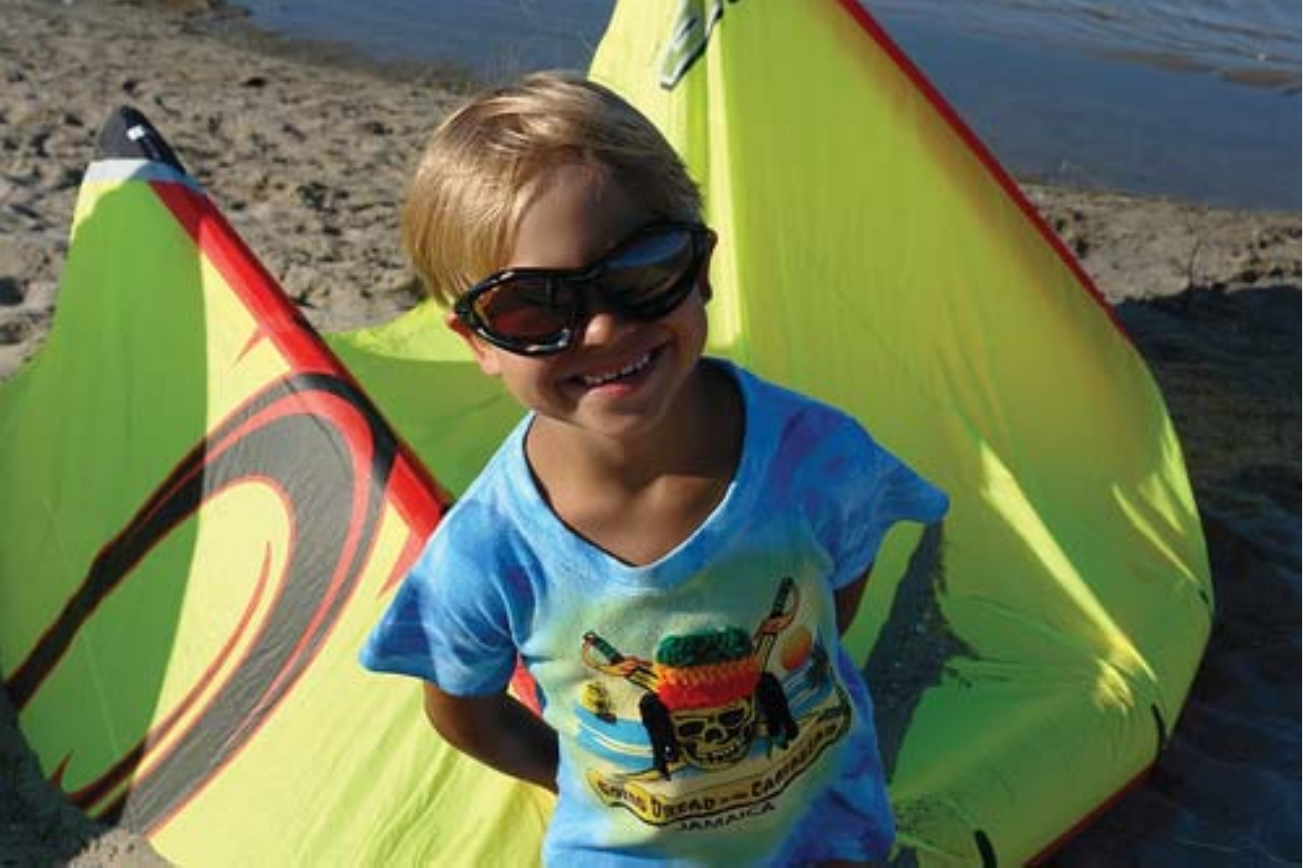}}}   & \\
& A young man wearing swimming goggles wearing a blue shirt with a pirate skull on it.    \\
& BoW$\rightarrow$422  \qquad  word2vec$\rightarrow$105 \qquad  \textbf{GRU$\rightarrow$7}\\
& \\ [4pt]
\multirow{4}{*}{\raisebox{-0.5\totalheight}{\includegraphics[width=1.8cm,height=1.8cm]{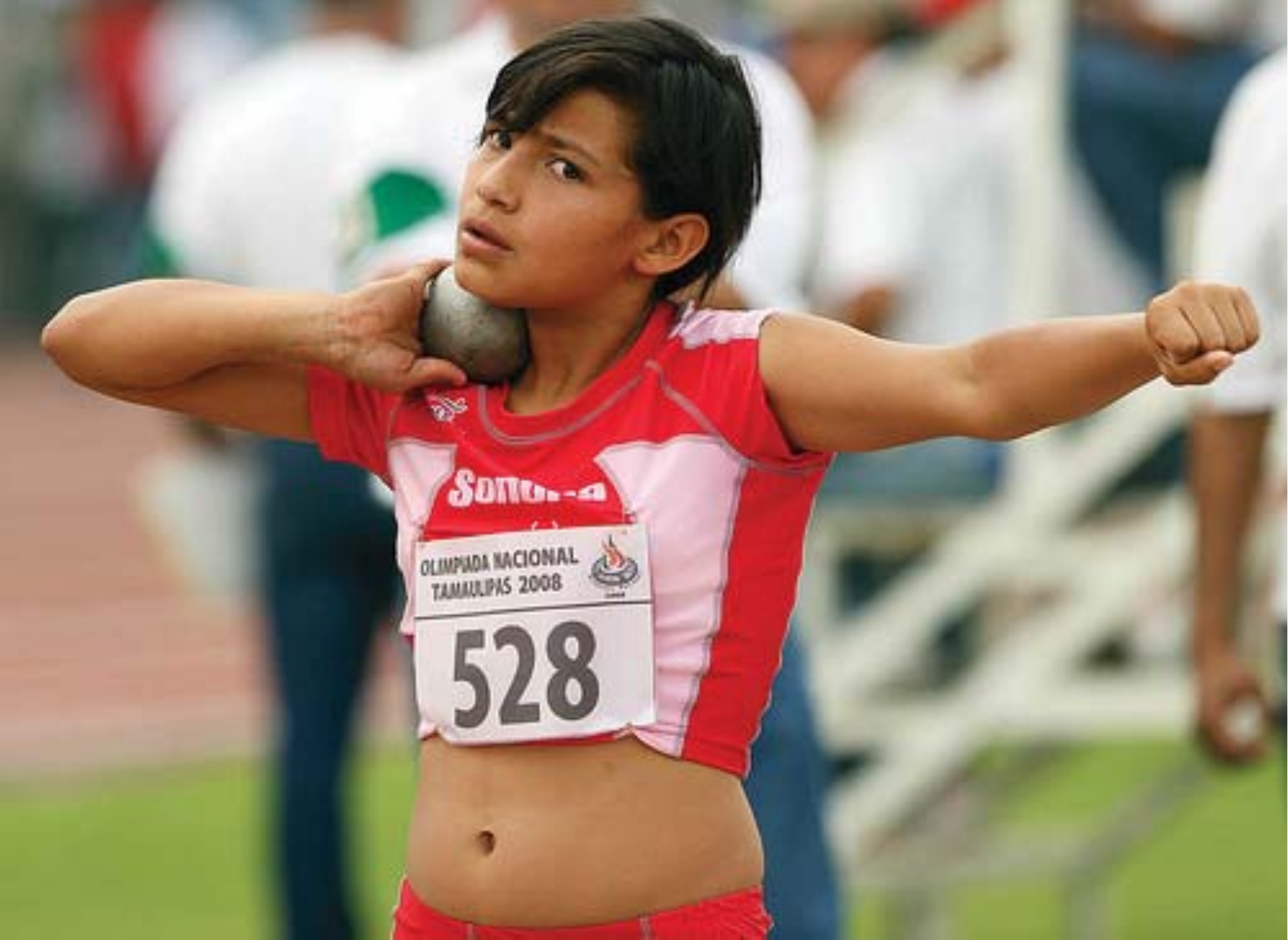}}}   & \\
& A dark-haired young woman, number 528, wearing red and white, is preparing to throw a shot put.    \\
& BoW$\rightarrow$80  \qquad  word2vec$\rightarrow$61 \qquad  \textbf{GRU$\rightarrow$1}\\
& \\
\bottomrule
\end{tabular}
 }
\end{table}

\textbf{Which visual feature?} 
%
Table \ref{tab:exp-flickr8k} and \ref{tab:exp-flickr30k} show performance of image caption retrieval on Flickr8k and Flickr30k, respectively. As the ConvNets go deeper, predicting the corresponding visual features by \ourmodel~improves. This result is encouraging as better performance can be expected from the continuous progress in deep learning features.
Table \ref{tab:exp-msvd} shows performance of video caption retrieval on MSVD, where the more compact GoogLeNet-shuffle feature tops the performance when combined with multi-scale sentence vectorization.
Although MSVD has more visual / sentence pairs than Flickr8k, it has a much less number of 1,200 visual examples for training.
Substituting ResNet-152 for GoogLeNet-shuffle reduces the amount of trainable parameters by 18\%, making Word2VisualVec more effective to learn from relatively limited examples. 
Ideally, the learning process shall allow the model to automatically discover which elements in the composite sentence vectorization layer are the most important for the problem in consideration. This advantage cannot be properly leveraged when training examples are in short supply. In such a case,  using word2vec instead of the composite vectorization is preferred, resulting in a Word2VisualVec with 73\% less parameters when using ResNet-152 (60\% less parameters when using CaffeNet or C3D) and thus easier to train. A similar phenomenon is observed on the image data, when given only 3k image-sentence pairs for training (see Fig. \ref{fig:increase-pairs}). Word2VisualVec with word2vec is more suited for small-scale training data regimes.

Given a fixed amount of training pairs, having more visual examples might be better for Word2VisualVec. To verify this conjecture, we take from the Flickr30k training set a random subset of 3k images with one sentence per image. We then incrementally increase the amount of image / sentence pairs for training, using the following two strategies.
One is to increase the number of sentences per image from 1 to 2, 3, 4, and 5 with the number of images fixed, while the other is to let the amount of images increase to 6k, 9k, 12k and 15k with the number of sentences per image fixed to one. As the performance curves in Fig. \ref{fig:increase-pairs} show, given the same amount of training pairs, adding more images results in better models. The result is also instructive for more effective acquisition of training data for image and video caption retrieval.

\begin{figure}[tb!]
\centering
\includegraphics[width=0.95\columnwidth]{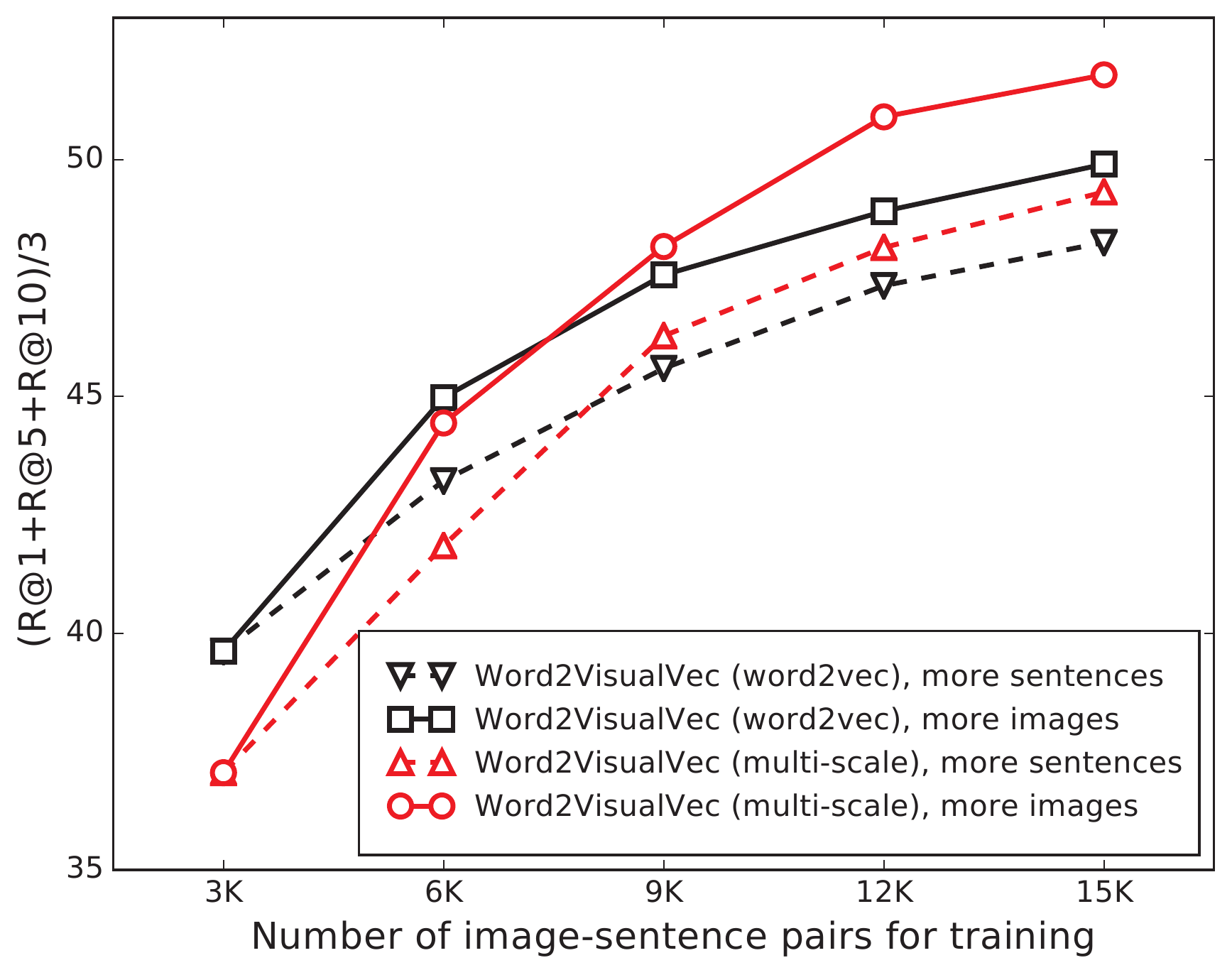}
\caption{Performance curves of two Word2VisualVec models on the Flickr30k test set, as the amount of image-sentence pairs for training increases. For both models, adding more training images gives better performance compared to adding more training sentences.}
\label{fig:increase-pairs}
\end{figure}

\textbf{How deep?} 
In this experiment, we use \wordvec~as sentence vectorization for its efficient execution. We vary the number of MLP layers, and observe a performance peak when using three-layers, \ie 500-2048-2048, on Flickr8k and four-layers, \ie 500-2048-2048-2048, on Flickr30k. Recall that the model is chosen in terms of its performance on the validation set. While its learning capacity increases as the model goes deeper, the chance of overfitting also increases.  To improve generalization we also tried $l_2$ regularization on the network weights. This tactic brings a marginal improvement, yet introduces extra hyper parameters. So we did not go further in that direction. Overall the three-layer \ourmodel~strikes the best balance between model capacity and generalization ability, so we use this network configuration in what follows.

\textbf{How fast?}
We implement \ourmodel~using Keras with theano backend. The three-layer model with multi-scale sentence vectorization takes about 1.3 hours to learn from the 30k image-sentence pairs in Flickr8k on a GeoForce GTX 1070 GPU. Predicting visual features for a given sentence is swift, at an averaged speed of 20 milliseconds. Retrieving captions from a pool of 5k sentences takes 8 milliseconds per test image. 
Based on the above evaluations we recommend \ourmodel~that uses multi-scale sentence vectorization, and predicts the 2,048-dim ResNet-152 feature when adequate training data is available (over 2k training images with five sentences per image) or the 1,024-dim GoogLeNet-shuffle feature when training data is more scarce.

\begin{figure*}[tb!]
\centering
\includegraphics[width=2.05\columnwidth]{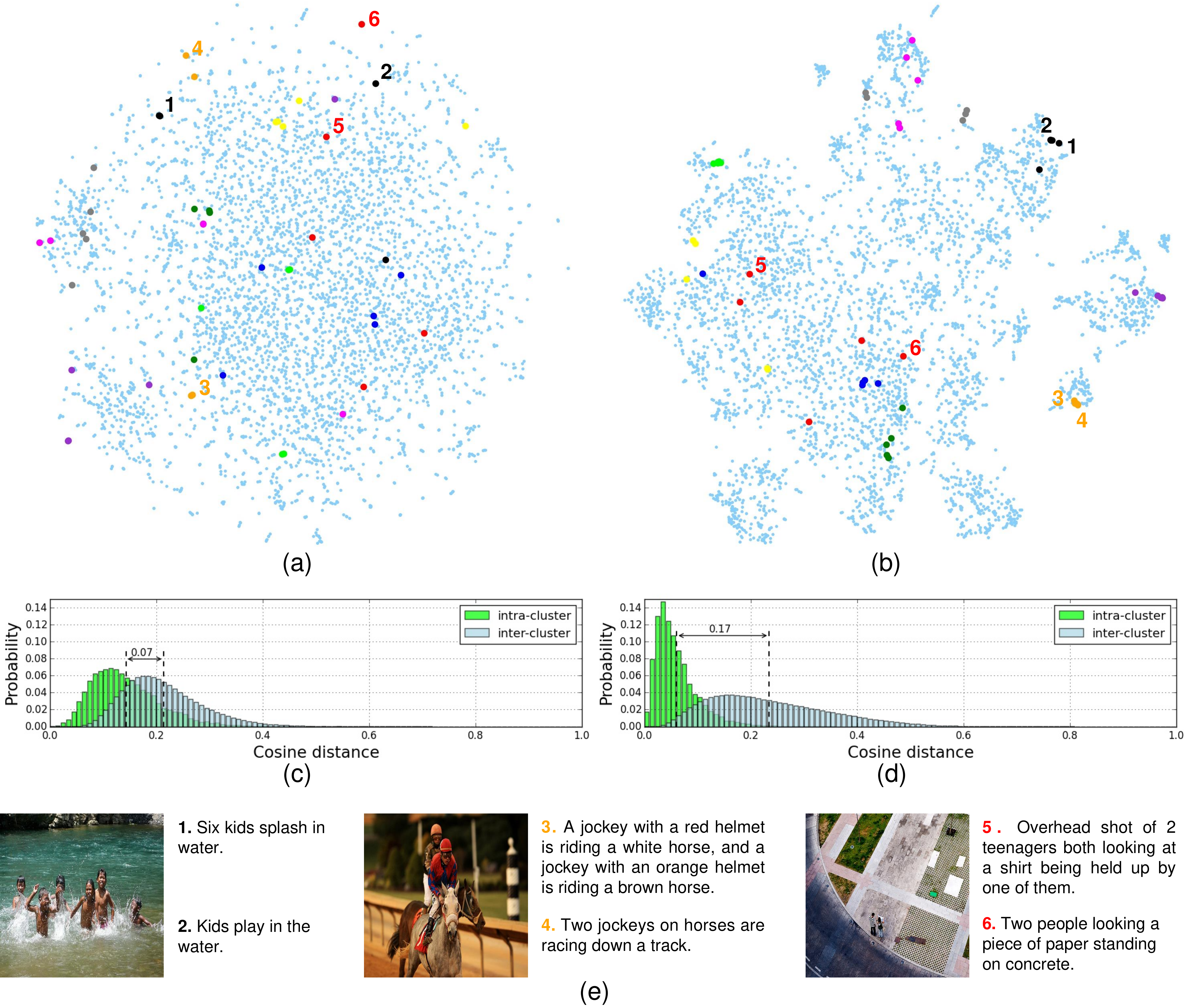}
\caption{
\textbf{\ourmodel~versus word2vec}.
For the 5k test sentences from Flickr30k, we use t-SNE \cite{maaten2008tsne} to visualize their distribution in (a) the word2vec space and (b) the Word2VisualVec space obtained by mapping the word2vec vectors to the ResNet-152 features. Histograms of intra-cluster (\ie sentences describing the same image) and inter-cluster (\ie sentences from different images) distances in the two spaces are given in (c) and (d). Bigger colored dots indicate 50 sentences associated with 10 randomly chosen images, with exemplars detailed in (e). Together, the plots reveal that different sentences describing the same image stay closer, while sentences from different images are more distant in the \ourmodel~space. Best viewed in color.
}\label{fig:tsne}
\end{figure*}


\begin{figure*}[tb!]
    \centering
        \includegraphics[width=0.81\linewidth]{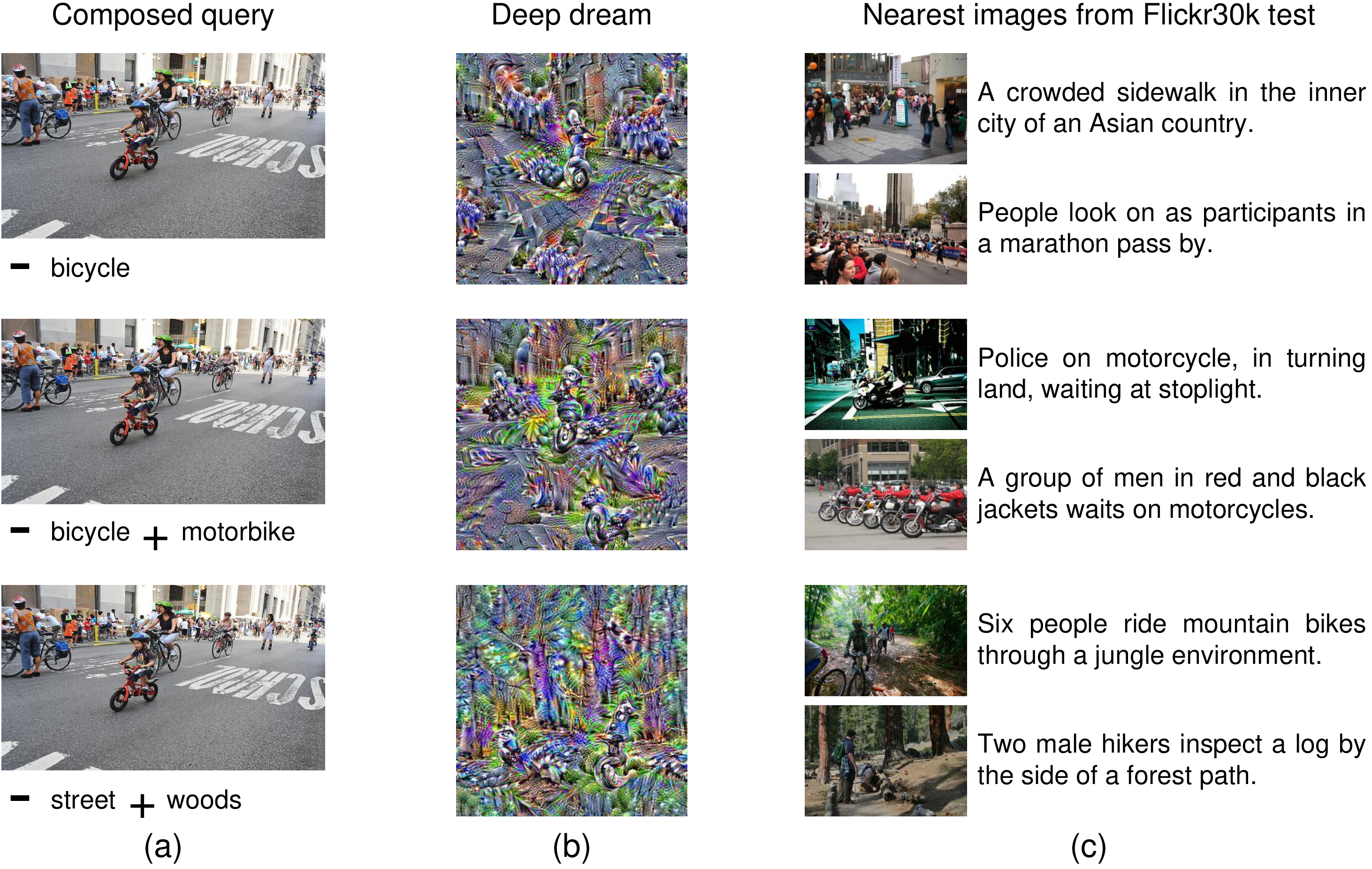}
    \caption{\textbf{\ourmodel~allows for multi-modal query composition.} (a) For each multi-modal query we visualize its predicted visual feature in (b) and show in (c) the nearest images and their sentences from the Flickr30k test set. Note the change in emphasis in (b), better viewed digitally in close-up.}
    \label{fig:visual}
\end{figure*}

\subsection{\ourmodel~versus word2vec}
Although our model is meant for caption retrieval, it essentially generates a new representation of text. How meaningful is this new representation as compared to word2vec?
To answer this question, we take all the 5K test sentences from Flickr30k, 
vectorizing them by word2vec and \ourmodel, respectively.
The word2vec model was trained on Flickr tags as described in Section \ref{sssec:arch}.
For a fair comparison, we let \ourmodel~use the same word2vec as its first layer.
Fig. \ref{fig:tsne} presents t-SNE visualizations of sentence distributions in the word2vec and \ourmodel~spaces,
showing that sentences describing the same image stay more close while sentences from distinct images are more distant in the latter space.
Recall that sentences associated with the same image are meant for describing the same visual content. Moreover, since they were independently written by distinct users, the wording may vary across the users, requiring a text representation to capture shared semantics among distinct words. 
\ourmodel~better handles such variance in captions as illustrated in the first two examples in Fig. \ref{fig:tsne}(e). 

The last example in Fig. \ref{fig:tsne}(e) shows failures of both models, where the two sentences (\#5 and \#6) are supposed to be close. Large difference between their subject (\textit{teenagers} versus \textit{people}) and object (\textit{shirt} versus \textit{paper}) makes it difficult for \ourmodel~to predict similar visual features from the two sentences. Actually, we find in the \ourmodel~space that the sentence nearest to \#5 is ``A woman is completing a picture of a young woman'' (which resembles subjects, \ie \emph{teenager} versus \emph{young woman} and action, \ie \emph{holding paper or easel}) and the one to \#6 is ``Kids scale a wall as two other people watch'' (which depicts similar subjects, \ie \emph{two people} and objects, \ie \emph{concrete} versus \emph{wall}).
This example shows the existence of large divergence between manually written descriptions of the same visual content, and thus the challenging nature of the caption retrieval problem.

Note that the above comparison is not completely fair as word2vec is not intended for fitting the relevance between image and text. By contrast, Word2VisualVec is designed to exploit the link between the two modalities, producing a new representation of text that is well suited for image and video caption retrieval.

\subsection{\ourmodel~for multi-modal querying}
Fig. \ref{fig:visual} presents an example of \ourmodel's learned representation and its ability for multi-modal query composition.  Given the query image, its composed queries are obtained by subtracting and/or adding the visual features of the query words, as predicted by \ourmodel. A deep dream visualization \cite{deepdream} is performed on an average (gray) image guided by each composed query. Consider the query in the second row for instance, where we instruct the search to replace bicycle with motorbike via a textual specification. The predicted visual feature of word \emph{bicycle} is subtracted (effect visible in first row) and the predicted visual feature of word \emph{motorbike} is added. Imagery of motorbikes are indeed present in the dream. Hence, the nearest retrieved images emphasize on motorbikes in street scenes.

 \label{ssec:exp-w2vv-learnt}



\subsection{Comparison to the State-of-the-Art}  \label{ssec:image-to-text}


\textbf{Image caption retrieval}. 
We compare a number of recently developed models for image caption retrieval \cite{iccv15-huawei,cvpr2015-klein-fv,iccv15-flickr30kentities,cvpr2016-wang,eccv2016-rnn_fv,tacl2015-kiros,iclr2016-vendrov}. 
All the methods, including ours, require image-sentence pairs to train. They all perform caption retrieval on a provided set of test sentences.
Note that the compared methods have no reported performance on the ResNet-152 feature. We have tried the VGGNet feature as used in \cite{iccv15-huawei,cvpr2015-klein-fv,cvpr2016-wang} and found Word2VisualVec less effective. This is not surprising as the choice of the visual feature is an essential ingredient of our model. While it would be ideal to replicate all methods using the same ResNet feature, only \cite{tacl2015-kiros,iclr2016-vendrov} have released their source code.
%
So we re-train these two models with the same ResNet features we use. Table \ref{tab:image_to_sent_perf} presents the performance of the above models on both Flickr8k and Flickr30k. \ourmodel~compares favorably against the state-of-the-art.
Given the same visual feature, our model outperforms \cite{tacl2015-kiros,iclr2016-vendrov},  especially for $R@1$.}
Notice that Plummer \etal \cite{iccv15-flickr30kentities} employ extra bounding-box level annotations. Still our results are better, indicating that we can expect further gains by including locality in the Word2VisualVec representation. As all the competitor models use joint subspaces, the results justify the viability of directly using the deep visual feature space for image caption retrieval.

Compared with the two top-performing methods \cite{cvpr2016-wang,iclr2016-vendrov}, the run-time complexity of the multi-scale Word2VisualVec is $O(m\times s+s\times g+(m+s+g)\times 2048+2048\times d)$, where $s$ indicates the dimensionality of word embedding and $g$ denotes the size of GRU. This complexity is larger than \cite{iclr2016-vendrov} which has a complexity of $O(m\times s+s\times g+g\times d)$, but lower than \cite{cvpr2016-wang} which vectorizes a sentence by a time-consuming Fisher vector encoding. 

\begin{table} [tb!]
\renewcommand{\arraystretch}{1.2}
\caption{\textbf{State-of-the-art for image caption retrieval}.
All numbers are from the cited papers except for \cite{tacl2015-kiros,iclr2016-vendrov}, both re-trained using their code with the same ResNet features we use.
\ourmodel~outperforms recent alternatives.}
\label{tab:image_to_sent_perf}
\centering 
\scalebox{0.95}{
\begin{tabular}{@{}l  rrr r@{\hskip 0.25in}  rrr @{}}
\toprule

 & \multicolumn{3}{c}{\textbf{Flickr8k}} & & \multicolumn{3}{c}{\textbf{Flickr30k}} \\
\cmidrule{2-4}  \cmidrule{6-8}
  & R@1 & R@5 & R@10 &   & R@1 & R@5 & R@10 \\
\cmidrule{1-8}
Ma \etal \cite{iccv15-huawei}                   & 24.8 & 53.7 & 67.1     &   &  33.6 & 64.1 & 74.9  \\
Kiros \etal  \cite{tacl2015-kiros}              & 23.7 & 53.1 & 67.3     &   &  32.9 & 65.6 & 77.1  \\ 
Klein \etal \cite{cvpr2015-klein-fv}            & 31.0 & 59.3 & 73.7     &   &  35.0 & 62.0 & 73.8  \\
Lev \etal \cite{eccv2016-rnn_fv}                & 31.6 & 61.2 & 74.3     &   &  35.6 & 62.5 & 74.2  \\ 
Plummer \etal \cite{iccv15-flickr30kentities}   & -- & -- & --           &   &  39.1 & 64.8 & 76.4  \\
Wang \etal \cite{cvpr2016-wang}                 & -- & -- & --           &   &  40.3 & 68.9 & 79.9  \\
Vendrov \etal  \cite{iclr2016-vendrov}          & 27.5 & 56.5 & 69.2     &   &  41.3 & 71.0 & 80.8  \\
\ourmodel & \textbf{36.3} & \textbf{66.4} &  \textbf{78.2}  &   & \textbf{45.9} &  \textbf{71.9} &  \textbf{81.3}  \\
\bottomrule
\end{tabular}
 }
\end{table}

\textbf{Video caption retrieval}. 
We also participated in the NIST TrecVid 2016 video caption retrieval task~\cite{AwadTRECVID16}. The test set consists of 1,915 videos collected from Twitter Vine. Each video is about 6 sec long. The videos were given to 8 annotators to generate a total of 3,830 sentences, with each video associated with two sentences written by two different annotators. The sentences have been split into two equal-sized subsets, set $A$ and set $B$, with the rule that sentences describing the same video are not in the same subset. Per test video, participants are asked to rank all sentences in the two subsets. Notice that we have no access to the ground-truth, as the test set is used for blind testing by the organizers only. NIST also provides a training set of 200 videos, which we consider insufficient for training \videovec. Instead, we learn the network parameters using video-text pairs from MSR-VTT \cite{xu2016msr}, with hyper-parameters tuned on the provided TrecVid training set. By the time of TrecVid submission, we used GoogLeNet-shuffle as the visual feature, a 1,024-dim bag of MFCC as the audio feature, and \wordvec~for sentence vectorization. The performance metric is Mean Inverted Rank (MIR) at which the annotated item is found. 
Higher MIR means better performance.

\begin{figure}[tb!]
\centering
\includegraphics[width=0.9\columnwidth]{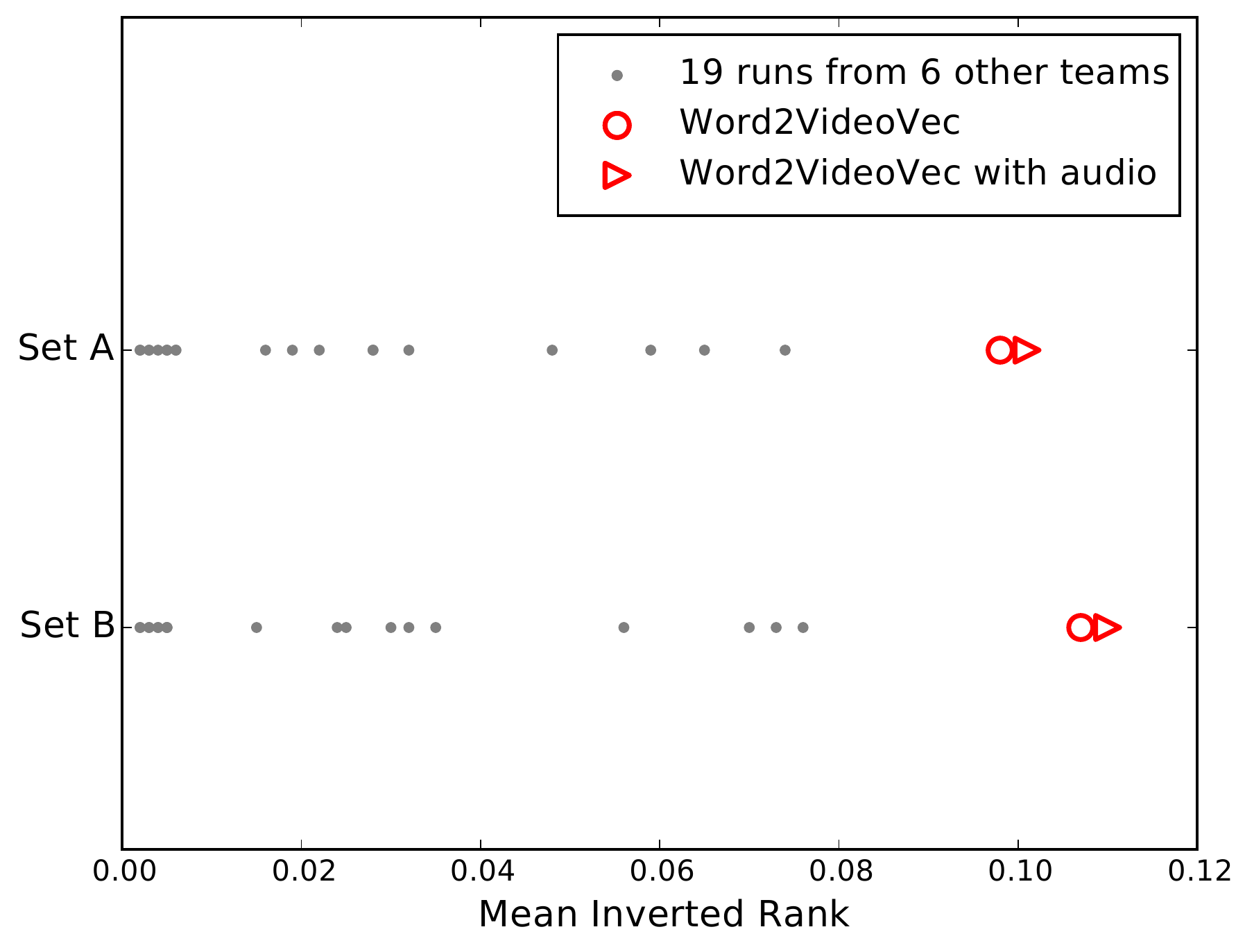}
\caption{\textbf{State-of-the-art for video caption retrieval} in the TrecVid 2016 benchmark, showing the good performance of \videovec~compared to 19 alternative approaches evaluated by the NIST TrecVid 2016 organizers \cite{AwadTRECVID16},
which can be further improved by predicting the visual-audio feature.}
\label{fig:trecvid_submission}
\end{figure}

As shown in Fig. \ref{fig:trecvid_submission},  with MIR ranging from 0.097 to 0.110, \videovec~leads the evaluation on both set A and set B in the context of 21 submissions from seven teams worldwide.
Moreover, the results can be further improved by predicting the visual-audio feature. 
Besides us two other teams submitted their technical reports, scoring their best MIR of 0.076  \cite{zhang2016vireo} and 0.006 \cite{le2016nii}, respectively. Given a video-sentence pair, the model from \cite{zhang2016vireo} iteratively combines the video and sentence features into one vector, followed by a fully connected layer to predict the similarity score. The model from \cite{le2016nii} learns an embedding space by minimizing a cross-media distance.
%

\begin{figure*}[tb!]
\centering\includegraphics[width=2\columnwidth]{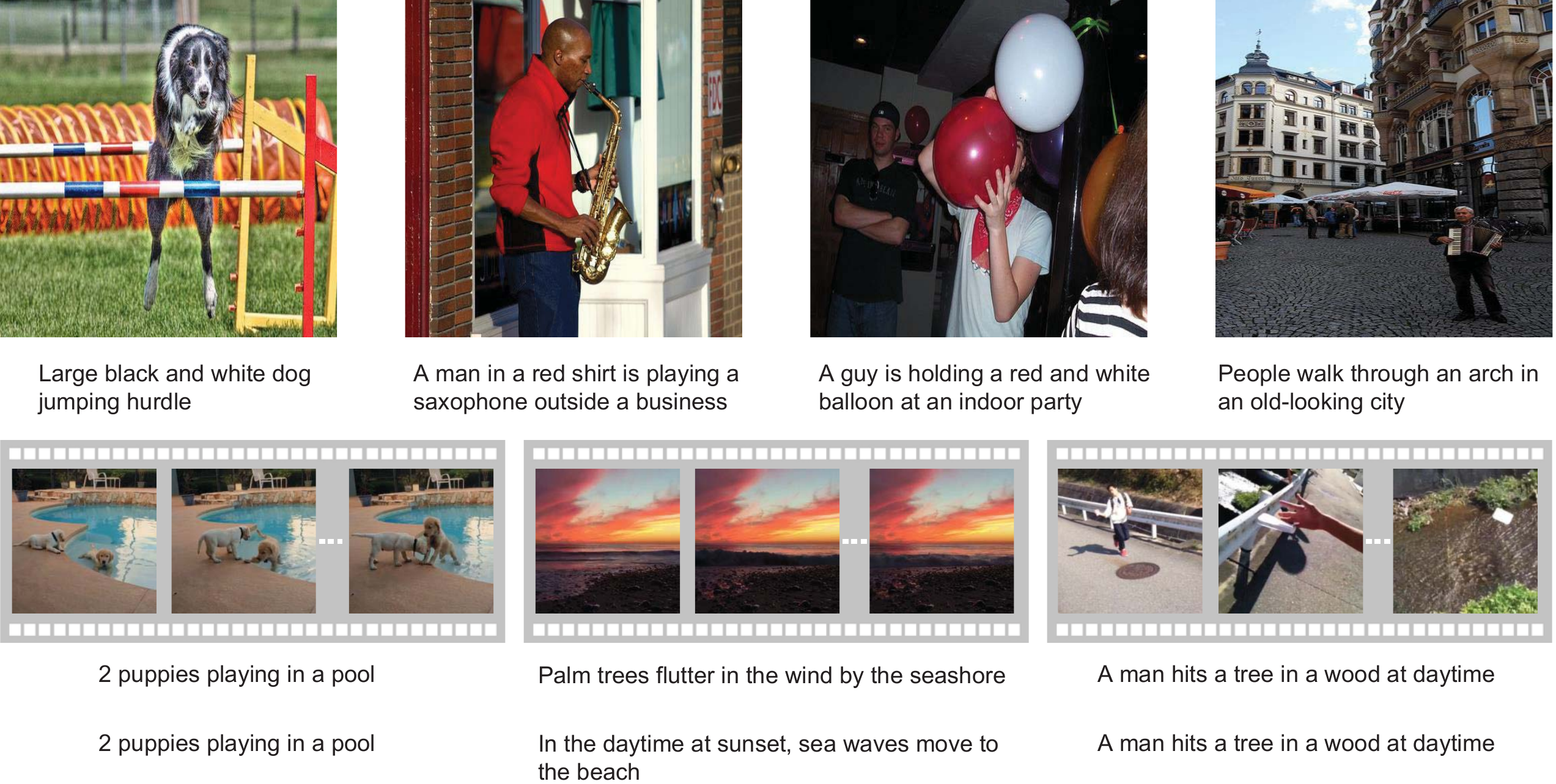}
\caption{Some image and video caption retrieval results by this work.
The last row are the sentences retrieved by \videovec~with audio, showing that adding audio sometimes help describe acoustics, e.g. \textit{sea wave} and \textit{speak}. 
}\label{fig:caption-results}
\end{figure*}

Some qualitative image and video caption retrieval results are shown in Fig. \ref{fig:caption-results}.
Consider the last image in the top row. Its ground-truth caption is ``A man playing an accordion in front of buildings'', while the top-retrieved caption is ``People walk through an arch in an old-looking city''. Though the ResNet feature well describes the overall scene, it fails to capture the \textit{accordion} which is small but has successfully drawn the attention of the annotator who wrote the ground-truth caption. 
The last video in the bottom row of Fig. \ref{fig:caption-results} shows ``A man throws his phone into a river''. This action is not well described by the averagely pooled video feature. Hence, the main sources of errors come from the cases where the visual features do not well represent the visual content.

\subsection{Limits of caption retrieval and possible extensions}  \label{ssec:limits}
The caption retrieval task works with the assumption that for a query image or video, there is at least one sentence relevant w.r.t the query. In a general scenario where the query is unconstrained with arbitrary content, this assumption is unlikely to be valid. A naive remedy would be to enlarge the sentence pool. A more advanced solution is to combine with methods that construct novel captions. 
In \cite{kuznetsova2012collective, ordonez2016large} for instance, a caption is formed using a set of visually relevant phrases extracted from a large-scale image collection. From the top-n sentences retrieved by Word2VisualVec, one can also generate a new caption, using the methods of \cite{kuznetsova2012collective, ordonez2016large}. As this paper is to retrieve rather than to construct a caption, we leave this for future exploration.

\section{Conclusions} \label{sec:conc}
This paper shows the viability of resolving image and video caption retrieval in a visual feature space exclusively. We contribute \ourmodel, which is capable of transforming a natural language sentence to a meaningful visual feature representation. Compared to the word2vec space, sentences describing the same image tend to stay closer, while sentences from different images are more distant in the \ourmodel~space.  As the sentences are meant for describing visual content, the new textual encoding captures both semantic and visual similarities. \ourmodel~also supports multi-modal query composition, by subtracting and/or adding the predicted visual features of specific words to a given query image. What is more the \ourmodel~ is easily generalized to predict a visual-audio representation from text for video caption retrieval. For state-of-the-art results,  we suggest \ourmodel~with multi-scale sentence vectorization, predicting the ResNet feature when adequate training data is available or the GoogLeNet-shuffle feature when training data is in short supply.  


\bibliographystyle{IEEEtran}
\bibliography{IEEEabrv,w2vv}

\begin{IEEEbiography}[{\includegraphics[width=1in,height=1.25in,clip,keepaspectratio]{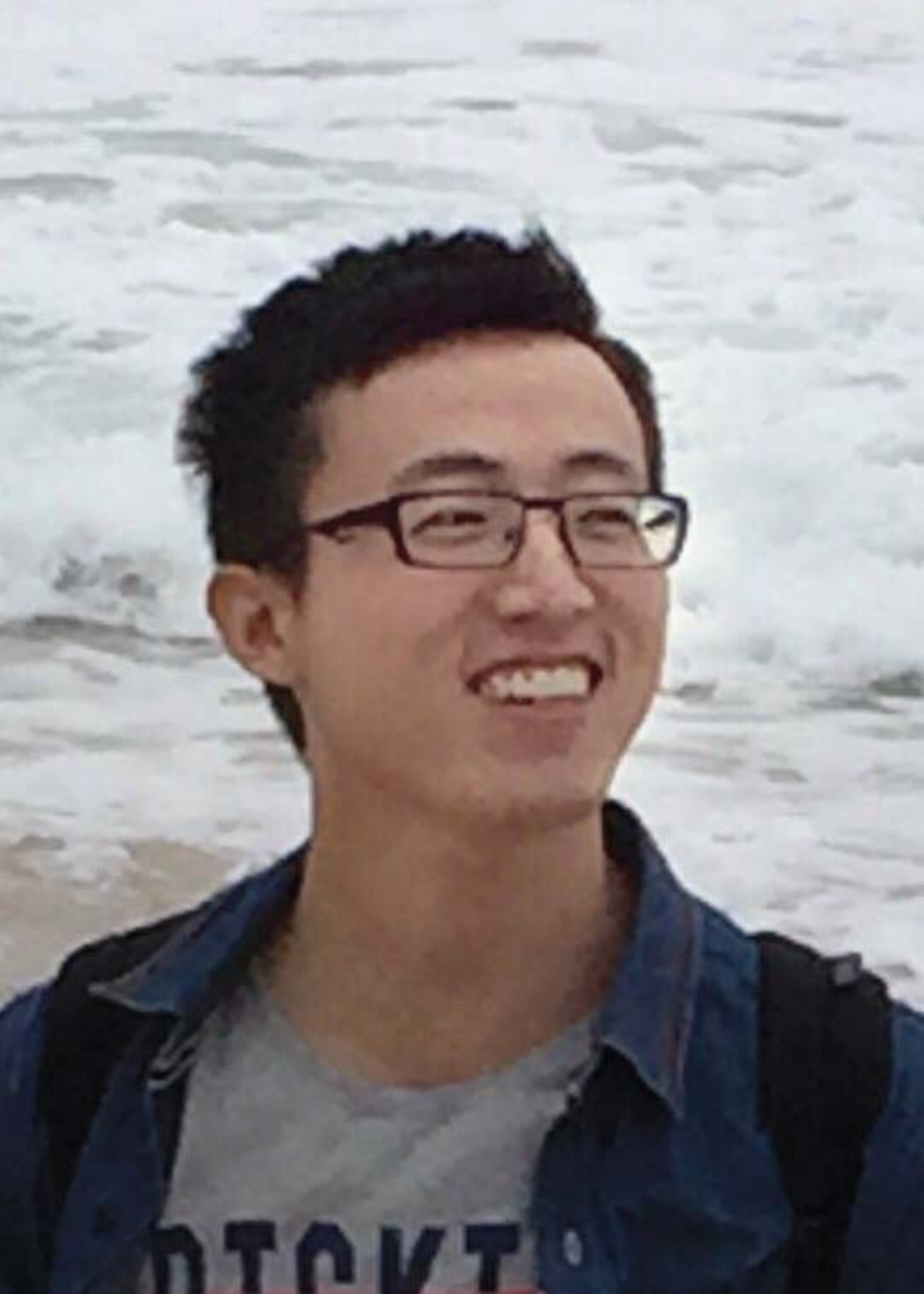}}]
{Jianfeng Dong} received the B.E. degree in software engineering from Zhejiang University of Technology, Hangzhou, China, in 2013.
He is currently a Ph.D. candidata in the School of Computer Science and Technology, Zhejiang University, Hangzhou, China.

His research interest is cross-media retrieval and deep learning. 
He was awarded the ACM Multimedia Grand Challenge Award in 2016.
\end{IEEEbiography}

\begin{IEEEbiography}[{\includegraphics[width=1in,height=1.25in,clip,keepaspectratio]{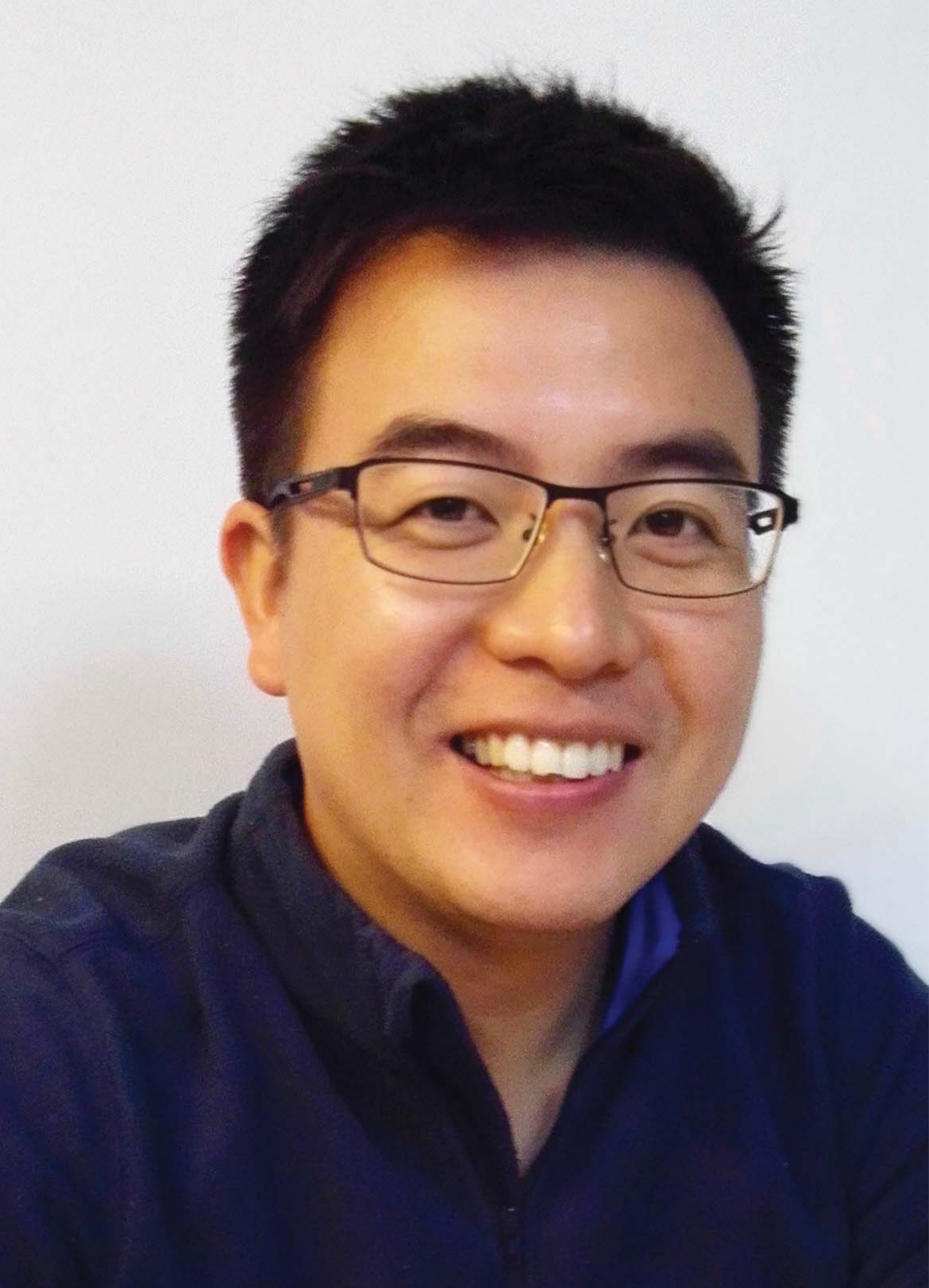}}]
{Xirong Li} received the B.S. and M.E. degrees from Tsinghua University, Beijing, China, in 2005 and 2007, respectively, and the Ph.D. degree from the University of Amsterdam, Amsterdam, The Netherlands, in 2012, all in computer science.

He is currently an Associate Professor with the Key Lab of Data Engineering and Knowledge Engineering, Renmin University of China, Beijing, China. His research includes image and video retrieval.

Prof. Li was an Area Chair of ICPR 2016 and Publication Co-Chair of ICMR 2015. He was the recipient of the ACM Multimedia 2016 Grand Challenge Award, the ACM SIGMM Best Ph.D. Thesis Award 2013, the IEEE TRANSACTIONS ON MULTIMEDIA Prize Paper Award 2012, the Best Paper Award of the ACM CIVR 2010, the Best Paper Runner-Up of PCM 2016 and PCM 2014 Outstanding Reviewer Award.

\end{IEEEbiography}


\begin{IEEEbiography}[{\includegraphics[width=1in,height=1.25in,clip,keepaspectratio]{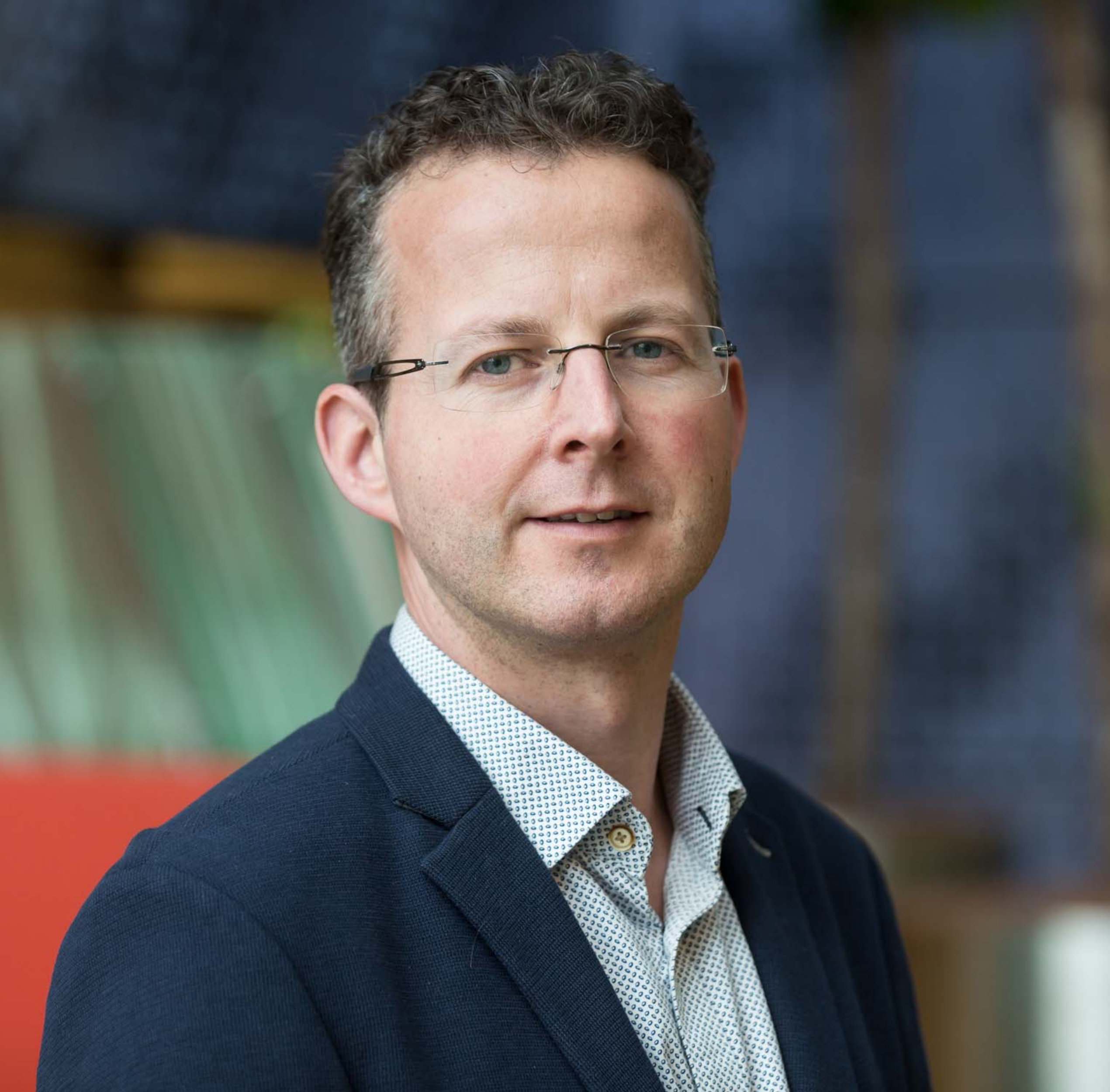}}]
{Cees G.M. Snoek} is a full professor in computer science at the University of Amsterdam, where he heads the Intelligent Sensory Information Systems Lab. He is also a director of the QUVA Lab, the joint research lab of Qualcomm and the University of Amsterdam on deep learning and computer vision. He received the M.Sc. degree in business information systems (2000) and the Ph.D. degree in computer science (2005) both from the University of Amsterdam, The Netherlands. He was previously an assistant and associate professor at the University of Amsterdam, as well as visiting scientist at Carnegie Mellon University and UC Berkeley, head of R\&D at University spin-off Euvision Technologies and managing principal engineer at Qualcomm Research Europe. His research interests focus on video and image recognition. He has published over 200 refereed journal and conference papers, and frequently serves as an area chair of the major conferences in multimedia and computer vision.

Professor Snoek is the lead researcher of the award-winning MediaMill Semantic Video Search Engine, which is the most consistent top performer in the yearly NIST TRECVID evaluations. He was general chair of ACM Multimedia 2016 in Amsterdam, founder of the VideOlympics 2007-2009 and a member of the editorial board for ACM Transactions on Multimedia. Cees is recipient of an NWO Veni award, a Fulbright Junior Scholarship, an NWO Vidi award, and the Netherlands Prize for ICT Research. Several of his Ph.D. students and Post-docs have won awards, including the IEEE Transactions on Multimedia Prize Paper Award, the SIGMM Best Ph.D. Thesis Award, the Best Paper Award of ACM Multimedia, an NWO Veni award and the Best Paper Award of ACM Multimedia Retrieval. Five of his former mentees serve as assistant and associate professors.

\end{IEEEbiography}



\end{document}